\documentclass[11pt,a4paper]{article}
\usepackage{note}
\usepackage{bm}
\usepackage[margin=1in]{geometry}
\usepackage{color}
\usepackage{cite}
\usepackage{algorithm}
\usepackage{algpseudocode}
\usepackage{subcaption}
\usepackage{graphicx}

\newtheorem{theorem}{\bf Theorem}[section]
\newtheorem{lemma}{\bf Lemma}[section]
\newtheorem{assumption}{\bf Assumption}[subsection]
\newtheorem{condition}{\bf}[subsection]

\date{}

\title{A Multi-Agent Off-Policy Actor-Critic Algorithm for Distributed
Reinforcement Learning}
\author{Wesley Suttle, Zhuoran Yang, Kaiqing Zhang, Zhaoran Wang, Tamer Ba\c{s}ar, and Ji Liu\thanks{%This research was supported in part by ONR MURI Grant N00014-16-1-2710, and in part by the Australian Research Council under grants DP-130103610 and DP-160104500, and Data61-CSIRO.
W. Suttle is with the Department of Applied Mathematics and Statistics at Stony Brook University (\texttt{wesley.suttle@stonybrook.edu}).
Z. Yang is with the Department of Operations Research and Financial Engineering at Princeton University
(\texttt{zy6@princeton.edu}).
K. Zhang and T. Ba\c{s}ar are with the Coordinated Science Laboratory at University of Illinois at Urbana-Champaign (\texttt{\{kzhang66,basar1\}@illinois.edu}), and their research was supported in part by the US Army Research Laboratory (ARL) Cooperative Agreement W911NF-17-2-0196.
Z. Wang is with the Department of Industrial Engineering and Management Sciences at Northwestern University
(\texttt{zhaoran.wang@northwestern.edu}).
J. Liu is with the Department of Electrical and Computer Engineering at Stony Brook University
(\texttt{ji.liu@stonybrook.edu}).
}
}

\begin{document}

\maketitle

%%%%%%%%%%%%%%%%%%%%%%%%%%%%%%%%%%%%%%%%%%%%%%%%%%%%%%%%%%%%%%%%%%%%%%%%%%%%%%%%
\begin{abstract}

This paper extends off-policy reinforcement learning to the multi-agent case in which a set of networked agents communicating with their neighbors according to a time-varying graph collaboratively evaluates and improves a target policy while following a distinct behavior policy. To this end, the paper develops a multi-agent version of emphatic temporal difference learning for off-policy policy evaluation, and proves convergence under linear function approximation.
The paper then leverages this result, in conjunction with a novel multi-agent off-policy policy gradient theorem and recent work in both multi-agent on-policy and single-agent off-policy actor-critic methods, to develop and give convergence guarantees for a new multi-agent off-policy actor-critic algorithm.

\end{abstract}

%%%%%%%%%%%%%%%%%%%%%%%%%%%%%%%%%%%%%%%%%%%%%%%%%%%%%%%%%%%%%%%%%%%%%%%%%%%%%%%%

\section{Introduction} \label{intro}

The field of multi-agent reinforcement learning (MARL) has recently seen a flurry of interest in the control and broader machine learning communities. In this paper, we consider the distributed MARL setting, where a set of agents communicating via a connected but possibly time-varying communication network collaboratively perform policy evaluation or policy improvement while sharing only local information. Important recent theoretical works in this area include \cite{kar13}, where the communication network is incorporated into the underlying model, \cite{zhang, zhang18_2}, where the theoretical basis for distributed on-policy actor-critic methods is established, \cite{chen19, lin19}, where progress is made in developing communication-efficient algorithms for this setting, and \cite{doan19}, where key finite-time results for the multi-agent case are obtained. However, in order for methods based on these recent developments to find widespread future use in important potential application areas -- e.g. multi-player games, multi-robot motion planning, and distributed control of energy networks -- the development of theoretical tools enabling principled design of data- and resource-efficient algorithms is essential.

Off-policy reinforcement learning with importance sampling correction is an active research area that has recently been leveraged to develop data- and resource-efficient reinforcement learning algorithms for off-policy control. In such methods, an agent seeks to perform policy evaluation or policy improvement on a given target policy by generating experience according to a distinct behavior policy and reweighting the samples generated to correct for off-policy errors. Algorithms incorporating these methods include Retrace($\lambda$) and IMPALA \cite{munos16, espeholt18}, which use importance sampling to enable theoretically sound reuse of past experience and more intelligent use of multi-processing and parallel computing capabilities. The theory underlying such importance sampling-based off-policy methods is relatively well-developed. An off-policy extension of the well-known temporal differences (TD($\lambda$)) algorithm for policy evaluation \cite{sutton00}, called the method of emphatic temporal differences or ETD($\lambda$), has been developed and shown to converge under linear function approximation \cite{yu15, sutton16}. Following the foundational policy gradient theorem of \cite{sutton00} for the on-policy case, recent efforts in the area of off-policy policy improvement include \cite{silver, lillicrap, gu17-1, gu17-2}, as well as \cite{maie}, which builds off the off-policy policy gradient theorem of \cite{degris12} in the tabular case to prove convergence of the actor step under linear approximation architectures. None of these works, however, extends the off-policy policy gradient theorem to general continuously differentiable approximation architectures. To this end, and building on the off-policy policy evaluation results in \cite{yu15} and \cite{sutton16}, \cite{imani} proves an off-policy policy gradient theorem using the emphatic weightings that are central to ETD($\lambda$), and describes an off-policy actor-critic algorithm based on their result.

Given the usefulness of off-policy methods in the development of data- and resource-efficient algorithms for reinforcement learning, it is clear that extending such methods to distributed MARL is essential, since such methods can be used to help mitigate slower convergence rates inherited from the distributed setting. In this paper, we present a new off-policy actor-critic algorithm for distributed MARL and provide convergence guarantees. Our algorithm uses a novel multi-agent consensus-based version of ETD$(\lambda)$ for the critic updates and relies on a new multi-agent off-policy policy gradient theorem using emphatic weightings to enable each agent to compute its portion of the policy gradient for the actor updates. We furthermore prove that, under linear approximation architectures for the value function, the algorithm converges with probability one. Finally, we provide empirical results validating our theoretical convergence guarantees in the linear function approximation setting.

Though the area of off-policy actor-critic methods for distributed MARL is new, and the results we provide in this paper are novel, an additional recent work in off-policy actor-critic MARL \cite{yanzhang19} appeared within a few days of the first version of our current paper becoming available. In that work, a distributed off-policy actor-critic algorithm is proposed that uses consensus for the actor updates -- unlike \cite{zhang, zhang18_2} and our current work, which both use consensus for the critic step -- and gradient temporal differencing (GTD) \cite{lagoudakis, yu10, mahmood} for the critic updates, whereas the algorithm we propose uses ETD$(\lambda)$. The work and results in \cite{yanzhang19} are thus distinct and independent from our own, though they tackle the same important task of rigorously extending off-policy methods to the multi-agent setting. We note in passing that single-agent GTD methods for off-policy policy evaluation are quadratic in the approximator's number of parameters \cite{sutton16}, which has the potential to lessen, and even eliminate the complexity-reduction advantages of using function approximation. ETD methods, on the other hand, inherit the relative simplicity and linear complexity in the number of parameters of TD($\lambda$) \cite{yu15, sutton16}. Empirical comparison of our approach and that in \cite{yanzhang19} is definitely warranted, but we leave this task as a future direction. 

The paper is structured as follows. In Section \ref{model}, we provide a brief introduction to the multi-agent Markov decision process model, notations, and basic definitions and assumptions from reinforcement learning that we will need going forward. In Section \ref{etd}, we provide a discussion of ETD$(\lambda)$ that both gives an introduction to that algorithm and establishes important notation and concepts used in the multi-agent algorithm and its analysis. Next, we derive our multi-agent off-policy policy gradient theorem in Section \ref{maoppgt}. In Section \ref{algs}, we present both the single- and multi-agent versions of our off-policy policy gradient algorithm. In Section \ref{results:theoretical}, we present our main convergence results and clearly state the assumptions upon which they rely. We then provide empirical results to both back up our theory in Section \ref{results:empirical}. Finally, we give some concluding remarks and point to future directions in the conclusion.

\section{Model Formulation} \label{model}
The multi-agent reinforcement learning problem is formulated as a Markov decision process (MDP) model on a time-varying communication network, which is introduced in detail as follows.

Let $\mathcal{N} = \{ 1, \ldots, n\}$ denote a set of $n$ agents, and let $\{ G_t \}_{t \in \mathbb{N}} = \{ ( \mathcal{N}, \mathcal{E}_t ) \}_{t \in \mathbb{N}}$ denote a possibly time-varying sequence of connected, undirected graphs on $\mathcal{N}$, which depicts the neighbor relationships among the agents. Specifically, $(j,i)$ is an edge in $G_t$ whenever agents $j$ and $i$ can communicate. Then, $(S, A, P, \{r^i\}_{i \in \mathcal{N}}, \{G_t\}_{t \in \mathbb{N}}, \gamma)$ characterizes a networked multi-agent discounted MDP, where $S$ is the shared state space, $A = \prod_{i \in \mathcal{N}} A^i$ is the joint action space (which is assumed to be constant, and where $A^i$ is the action space of agent $i$), $P : S \times S \times A \rightarrow [0, 1]$ is the transition probability function, $r^i : S \times A \rightarrow [0, 1]$ is the local reward function for each agent $i \in \mathcal{N}$, the sequence $\{ G_t \}_{t \in \mathbb{N}}$ describes the communication network at each timestep, and $\gamma \in (0,1)$ is an appropriately chosen discount factor.

We assume that the state and action spaces are finite. We also assume that, for each graph $G_t$, there is an associated, nonnegative, possibly random weight matrix $C_t$ that respects the topology of $G_t$ in that, if $(i,j) \notin \mathcal{E}_t,$, then $[C_t]_{ij} = 0$. Several important assumptions about the sequence $\{C_t\}_{t \in \mathbb{N}}$ will be made explicit in Section \ref{assumptions} below. Finally, let $\bar{r}_{t+1}$ denote the global reward generated at time $t+1$, and let $\bar{r} : S \times A \rightarrow \mathbb{R}$ be given by $\bar{r}(s, a) = \frac{1}{n} \sum_{i \in \mathcal{N}} r^i(s, a) = E[\bar{r}_{t+1} \ | \ s_t = s, a_t = a]$.

Recall that a policy function $\nu : A \times S \rightarrow [0,1]$ leads to a conditional probability distribution $\nu(\cdot | s)$ over $A$ for each element $s \in S$. For a given policy $\nu$, the state-value function is 
%\small
$$v_{\nu}(s) = E_{s \sim \nu} \Big[ \sum_{k = 1}^{\infty} \gamma^{k-1} \bar{r}_{t+k} \ | \ s_t = s \Big],$$
%\normalsize
which satisfies 
%\small
$$v_{\nu}(s) = \sum_{a \in A} \nu(a|s) \sum_{s' \in S} P(s' | s, a) [\bar{r}(s, a) + \gamma v_{\nu}(s')].$$
%\normalsize
The action-value function is
%\small
$$q_{\nu}(s,a) = \sum_{s' \in S} P(s' |  s, a)(\bar{r}(s, a) + \gamma v_{\nu}(s')).$$
%\normalsize

Let each agent $i \in \mathcal{N}$ be equipped with its own local behavior policy $\mu^i : A^i \times S \rightarrow [0,1]$. For each $i \in \mathcal{N}$, let $\pi^i_{\theta^i} : A^i \times S \rightarrow [0, 1]$ be some suitable set of local target policy functions parametrized by $\theta^i \in \Theta^i$, where $\Theta^i \subset \rn{m_i}$ is compact. We further assume that each $\pi^i_{\theta^i}$ is continuously differentiable with respect to $\theta^i$. Set $\theta = [\theta_1^T, \ldots, \theta_n^T]^T$. Define 
\small
$$
\mu = \prod_{i=1}^n \mu^i : A \times S \rightarrow [0,1] \text{ and } \pi_{\theta} = \prod_{i=1}^n \pi^i_{\theta^i} : A \times S \rightarrow [0,1].
$$
\normalsize
These correspond to the global behavior function and global parametrized target policy function, respectively.

Assume that $\mu^i(a^i | s) > 0$ whenever $\pi^i_{\theta^i} (a^i | s) > 0$, for all $i \in \mathcal{N}$, all $(a^i, s) \in A^i \times S$, and all $\theta^i \in \Theta^i$. For all $\theta \in \Theta,$ assume that the Markov chains generated by $\pi_{\theta}$ and $\mu$ are irreducible and aperiodic, and let $\mathbf{d}_{\pi_{\theta}}, \mathbf{d}_{\mu} \in [0,1]^{|S|}$ denote their respective steady-state distributions, i.e. $d_{\pi_{\theta}}(s)$ is the steady-state probability of the $\pi_{\theta}$-induced chain being in state $s \in S$, and similarly for $d_{\mu}(s)$.

Finally, let each agent be equipped with a state value function estimator $v_{\omega^i} : S \rightarrow \r$ parametrized by $\omega^i \in \Omega$, where $\Omega \subset \rn{M}, M \in \mathbb{N}, M > 0$ is parameter space shared by all agents. This family of functions will be used in the following to maintain a running approximation of the true value function for the current policy. We emphasize that each agent maintains its own local estimate $\omega^i$ of the current value function parameters, but that all agents use identical approximation architectures, i.e. $v_{\omega^i} = v_{\omega^j}$ whenever $\omega^i = \omega^j$.  In the case of general approximation architectures, it is only required that $v_{\omega}$ be a suitably expressive approximator that is differentiable in $\omega$, such as a neural network. In our convergence analysis, however, we assume the standard linear approximation architecture $v_{\omega}(s) = \phi(s)^T \omega$, where $\phi(s)$ is the feature vector corresponding to $s \in S$.

\section{Emphatic Temporal Difference Learning} \label{etd}
Since we extend the single-agent ETD($\lambda$) algorithm developed in \cite{sutton16, yu15} to the multi-agent setting and use it to perform off-policy policy evaluation during the faster-timescale critic step of our algorithm, and since we also repurpose the emphatic weightings used in ETD$(\lambda)$ in the next section for use in the policy gradient theorem as in \cite{imani}, it is helpful to summarize the basic form of single-agent ETD($\lambda$) with linear function approximation in this section.

We are given a discounted MDP $(S, A, P, r, \gamma)$, target policy $\pi : A \times S \rightarrow [0,1]$, and behavior policy $\mu : A \times S \rightarrow [0,1]$, with $\pi \neq \mu$. It is assumed that the steady-state distributions $\bm{d}_{\pi}, \bm{d}_{\mu}$ of $\pi, \mu$ exist, and that the transition probability matrices that they induce are given by $P_{\pi}, P_{\mu}$. The goal is to perform on-line policy evaluation on $\pi$ while behaving according to $\mu$ over the course of a single, infinitely long trajectory. This is accomplished by carrying out TD($\lambda$)-like updates that incorporate importance sampling ratios to reweight the updates sampled from $\mu$ to correspond to samples obtained from $\pi$. At a given state-action pair $(s,a)$, the corresponding importance sampling ratio is given by $\rho(s,a) = \frac{\pi(a|s)}{\mu(a|s)}$, with the assumption that if $\pi(a|s) > 0$, then $\mu(a|s) > 0$, and the convention that $\rho(s,a) = 0$ if $\mu(a|s) = \pi(a|s) = 0$.

The work \cite{yu15} proves the convergence of ETD($\lambda$) with linear function approximation using rather general forms of discounting, bootstrapping, and a notion of state-dependent ``interest''. First, instead of a fixed discount rate $\gamma \in (0,1)$, a state-dependent discounting function $\gamma : S \rightarrow [0,1]$ is used. Second, a state-dependent bootstrapping parameter $\lambda : S \rightarrow [0,1]$ at each step is allowed. Finally, \cite{yu15} include an interest function $i : S \rightarrow \r_+$ that stipulates the user-specified interest in each state.

Let $\Phi \in \rn{|S| \times k}$ be the matrix whose rows are the feature vectors corresponding to each state in $S$, and let $\phi(s)$ denote the row corresponding to state $s$. Given a trajectory $\{(s_t, a_t)\}_{t \in \mathbb{N}}$, let $\phi_t = \phi(s_t), \rho_t = \rho(s_t,a_t), \gamma_t = \gamma(s_t), \lambda_t = \lambda(s_t),$ and $r_t = r(s_t,a_t)$. An iteration of the general form of ETD($\lambda$) using linear function approximation is as follows:
$$\omega_{t+1} = \omega_t + \alpha_t \rho_t e_t (r_{t+1} + \gamma_{t+1} \phi_{t+1}^T \omega_t - \phi_t^T \omega_t),$$
where the eligibility trace $e_t$ is defined by
$e_t = \lambda_t \gamma_t \rho_{t-1} e_{t-1} + M_t \phi_t$, and $M_t$ is the emphatic weighting given by 
$$M_t = \lambda_t i(s_t) + (1 - \lambda_t) F_t, \qquad F_t = \gamma_t \rho_{t-1} F_{t-1} + i(s_t).$$
The stepsizes
$\sq{\alpha_t}{t \in \n}$ satisfy the standard conditions $\alpha_t \geq 0, \sum_t \alpha_t = \infty, \sum_t \alpha^2_t < \infty$, and $(e_0, F_0, \omega_0)$ are specified initial conditions, which may be arbitrary.

The actual derivation of this algorithm would take us much too far afield, and we refer the reader to \cite{sutton16} for an intuitive description and complete derivation of ETD($\lambda$). It is important for our purposes, however, to recognize the projected Bellman equation that it almost surely (a.s.) solves, as well as the associated ordinary differential equation (ODE) that it asymptotically tracks a.s. In the following description, we rely heavily on \cite{yu15}. We will also need two important results regarding the trace iterates $\{ e_t \}_{t \in \mathbb{N}}$, which can be found in Section \ref{etd:traces} in the appendix.

Let $S = \{s_1, \ldots, s_k\}$ be an enumeration of $S$. Define diagonal matrices $\Gamma = \text{diag}(\gamma(s_1), \ldots, \gamma(s_k))$ and $\Lambda = \text{diag}(\lambda(s_1), \ldots, \lambda(s_k))$. Let $r_{\pi} \in \rn{k}$ be such that its $j$-th entry is given by $r(s_j,\pi(s_j))$, and define
$$P^{\lambda}_{\pi,\gamma} = I - (I - P_{\pi} \Gamma \Lambda)^{-1}(I - P_{\pi} \Gamma), \hspace{2mm} r^{\lambda}_{\pi,\gamma} = (I - P_{\pi} \Gamma \Lambda)^{-1} r_{\pi}.$$
Associated with ETD($\lambda$) is the generalized Bellman equation \cite{sutton95, yu15}
$$v = r^{\lambda}_{\pi,\gamma} + P^{\lambda}_{\pi,\gamma}v,$$
with unique solution which we denote by $v_{\pi}$. ETD($\lambda$) solves the projected Bellman equation
\eqn{etd:pbe1}{v = \Pi(r^{\lambda}_{\pi,\gamma} + P^{\lambda}_{\pi,\gamma} v),}
where $v$ is constrained to lie in the column space of $\Phi$, and $\Pi$ is the projection onto $\text{colsp}(\Phi)$ with respect to the Euclidean norm weighted by the diagonal matrix
$$\overline{M} = \text{diag}(\bm{d}^T_{\mu,i}(I - P^{\lambda}_{\pi,\gamma})^{-1}),$$
where $\bm{d}_{\mu,i}(s_j) = \bm{d}_{\mu}(s_j) \cdot i(s_j)$, for $j = 1, \ldots, k$. It does this by finding the solution to the equation
\eqn{etd:eq1}{C \omega + b = 0,}
where $\omega \in \rn{k}$ is the element in the approximation space $\rn{k}$ corresponding to the linear combination $\Phi \omega \in \text{colsp}(\Phi)$, and $C$ and $b$ are given by
$$C = - \Phi^T \overline{M}(I - P^{\lambda}_{\pi,\gamma}) \Phi, \hspace{2mm} b = \Phi^T \overline{M} r^{\lambda}_{\pi,\gamma}.$$
When $C$ is negative definite, ETD($\lambda$) is proven in \cite{yu15} to almost surely find the unique solution $\omega^{*} = -C^{-1}b$ of equation (\ref{etd:eq1}) above, which is equivalent to finding the unique element $\Phi \omega^{*} \in \text{colsp}(\Phi)$ solving (\ref{etd:pbe1}).

In our extension of ETD($\lambda$) to the multi-agent case, we make the notation-simplifying assumptions that $\gamma(s) = \gamma \in (0,1)$ and $\lambda(s) = \lambda \in [0,1],$ and $i(s) = 1$, for all $s \in S$.

\section{Multi-agent Off-policy Policy Gradient Theorem} \label{maoppgt}
Following \cite{degris12} and \cite{imani}, when performing gradient ascent on the global policy function, we seek to maximize
\eqn{off_objective}{J_{\mu}(\theta) = \sum_{s \in S} d_{\mu}(s) v_{\pi_{\theta}} (s).}
For an agent to perform its gradient update at each actor step, it needs access to an estimate of its portion of the policy gradient. In the single-agent case, \cite{imani} obtains the expression
$$\nabla_{\theta} J_{\mu}(\theta) = \sum_{s \in S} m(s) \sum_{a \in A} [\nabla_{\theta} \pi_{\theta} (a|s)] q_{\pi_{\theta}}(s,a),$$
for the policy gradient, where $m(s)$ is the emphatic weighting of $s \in S$, with vector form $\mathbf{m}^T = \mathbf{d}_{\mu}^T ( \mathbf{I} - \mathbf{P}_{\theta, \gamma})^{-1}$, where $\mathbf{P}_{\theta,\gamma} \in \mathbb{R}^{|S| \times |S|}$ has entries given by
$$\mathbf{P}_{\theta,\gamma}(s,s') = \gamma \sum_{a \in A} \pi_{\theta}(a|s)P(s' |  s,a).$$
Recall that $\theta^i$ is the parameter of the local target policy $\pi^i_{\theta^i}$, $\forall i\in \mathcal{N}$. We will henceforth use the shorthand $q_{\theta}$ to refer to the action-value function $q_{\pi_{\theta}}$ of policy $\pi_{\theta}$. Building on the work in \cite{imani} and \cite{zhang}, which themselves are built on \cite{sutton00}, for the multi-agent case we obtain the following expression for the off-policy policy gradient in the multi-agent case, the proof of which is provided in the appendix:

\begin{theorem} \label{pgt}
The gradient of 
 $J_{\mu} (\theta)$ defined in \eqref{off_objective}  with respect to each $\theta^i$ is 
\eqn{pgt:gradient}{\nabla_{\theta^i} J_{\mu}(\theta) = \sum_{s \in S} m(s) \sum_{a \in A}\pi_{\theta}(a|s) q_{\theta}(s,a) \nabla_{\theta^i} \log \pi_{\theta^i}(a^i|s).}
\end{theorem}
\noindent It is also possible to incorporate baselines similar to those in \cite{zhang} in this expression, and the derivations are similar to those in that paper.

Let $\rho_t, F_t$ be as in the previous section, and let $\delta^i_t = r^i_{t+1} + \gamma v_{\omega^i_t}(s_{t+1}) - v_{\omega^i_t}(s_t)$ denote the temporal difference of the actor update at agent $i$ at time $t$. For the actor portion of our algorithm, we need a slightly different emphatic weighting update than that in ETD($\lambda$), corresponding to the update used in \cite{imani}. Define
\eqnn{M^{\theta}_t = (1-\lambda^{\theta}_t) + \lambda^{\theta}F_t = 1 + \lambda^{\theta} \gamma \rho_{t-1} F_{t-1}.}
In the actor portion of our algorithm given in the next section, we will be sampling from the expectation
\eqn{pgt:eq1}{\Es{\mu}{\rho_t M^{\theta}_t \delta^i_t \nabla_{\theta^i} \log \pi^i_{\theta^i_t}(a_t|s_t)}}
and using it as an estimate of the policy gradient at each timestep. To see why sampling from (\ref{pgt:eq1}) might give us an estimate of the desired gradient, note that, for fixed $\theta$,
$$\sum_{a \in A} \pi_{\theta}(a|s) q_{\theta}(s,a) \nabla_{\theta^i} \log \pi^i_{\theta^i}(a^i|s) = \sum_{a \in A} \mu(a|s) \rho_{\theta}(s,a) q_{\theta}(s,a) \nabla_{\theta^i} \log \pi^i_{\theta^i}(a^i|s).$$
To justify this sampling procedure, it is also important to note that, given the true $q_{\theta_t}$ for policy $\pi_{\theta_t}$, such sampling leads to unbiased estimates, i.e.
\eqn{pgt:eq2}{\sum_{s \in S} m(s) \sum_{a \in A} q_{\theta_t}(s_t,a_t) \nabla_{\theta^i} \pi_{\theta^i_t}(a^i_t|s_t) = \Es{\mu}{\rho_t M^{\theta}_t \delta^i_t \nabla_{\theta^i} \log \pi^i_{\theta^i_t}(a_t|s_t)}.}
Proof of (\ref{pgt:eq2}) in the single-agent case can be found in \cite{imani}, and the multi-agent case is an immediate consequence.

\section{Algorithms} \label{algs}

\subsection{Single-agent Algorithm} \label{algs:single}
Before introducing our multi-agent algorithm, we first describe the single-agent version. Recall that this is a two-timescale off-policy actor-critic algorithm, where the critic updates are carried out at the faster timescale using ETD($\lambda$) as in \cite{sutton16}, while the actor updates are performed at the slower timescale using the emphatically-weighted updates as in the previous section. The form of the following algorithm is based on \cite{imani}, but we choose an explicit method for performing the $\omega$ updates.

Let $\omega \in \Omega \subset \rn{k}$ and $\theta \in \Theta \subset \rn{l}$ be the value function and policy function parameters, respectively. For now, we can simply take $\Omega = \rn{k}$ and $\Theta = \rn{l}$. We will impose conditions on them ($\Theta$, in particular) in the Assumptions section below.

The single-agent version of the algorithm is as follows. We first initialize the parameters by setting $\theta_0 = 0, \ \omega_0 = e_{-1} = 0, \ F_{-1} = 0, \ \rho_{-1} = 1.$\footnote{\cite{imani} suggests $\lambda^{\theta} = 0.9$ as a default value. We currently have no suggestions for $\lambda$.} 
In each iteration, we execute action $a_t \sim \mu(\cdot|s_t) $ and observe $r_{t+1}$ and $s_{t+1}$. We then update the emphatic weightings by
$$M_t = \lambda + (1-\lambda) F_t,\qquad M^{\theta}_t = 1 + \lambda^{\theta} \gamma \rho_{t-1} F_{t-1}, $$ with $F_t = 1 + \gamma \rho_{t-1} F_{t-1}$.
 Finally, the actor and critic parameters are updated   using the emphatic weightings:
 \begin{align*}
     \omega_{t+1}& = \omega_t + \beta_{\omega,t} \rho_t (r_{t+1} + \gamma v_{\omega_t}(s_{t+1}) - v_{\omega_t}(s_t)) e_t, \\
     \theta_{t+1} &= \theta_t + \beta_{\theta,t} \rho_t M^{\theta}_t \nabla_{\theta} \log \pi_{\theta_t} (a_t|s_t) \delta_t,
 \end{align*}
where $e_t$ is given by 
$e_t = \gamma \lambda e_{t-1} + M_t \nabla_{\omega} v_{\omega_t}(s_t)$, and 
 $\delta_t = r_{t+1} + \gamma v_{\omega_t}(s_{t+1}) - v_{\omega_t}(s_t)$ is the standard TD(0) error. It is important to mention that $\delta_t$ can also be regarded as an estimate of the advantage function $q_{\pi}(s_t,a_t) - v_{\pi}(s_t)$, which is the standard example of including baselines. 

\subsection{Multi-agent Algorithm} \label{algs:multi}
With the above as a reference and jumping-off point, we are now ready to introduce our multi-agent off-policy actor-critic algorithm. The overall structure of the algorithm is similar to the single-agent version, with two key differences: (i) the agents perform the critic updates at the faster timescale using one consensus process to average their current $\omega$ estimates and an inner consensus process to obtain the importance sampling ratios necessary to perform ETD$(\lambda)$ for the current ``static'' global policy; (ii) each agent is responsible for updating only its own portion of the policy gradient at each actor update at the slower timescale.

All agents are initialized as in the single-agent case. At each step, each agent first performs a consensus average of its neighbor's $\omega$-estimates, selects its next action, and computes its local importance sampling ratio. Specifically, at the $t$-th iteration, 
agent $i$ first receives $ \widetilde{\omega}^j_{t-1}$ from each of its neighbors $j \in \mathcal{N}_t(i) $, executes its own action 
$a^i_t \sim \mu_i(\cdot|s_t)$, and observes the joint action $a_t$, its own  reward $r_{t+1}^i$, and the next state $s_{t+1}$. Agent $i$ then aggregates the information obtained from its neighbors with the consensus update
$ 
\omega^i_t = \sum_{j \in \mathcal{N}} c_{t-1}(i,j) \widetilde{\omega}^j_{t-1} 
$, and also computes the log of its local  importance sampling ratio 
$$
p_t^i = \log \bigl  [  \pi^i_{\theta^i_t}(a^i_t|s_t) \big / \mu_i(a^i_t|s_t) \bigr ].  $$
Here $c_t(i,j)$ is the communication weight from agents $j$ to $i$ at time $t$.
For undirected graphs, one particular choice of the weights $c_t(i,j)$ that relies on only local information of the agents is known as the Metropolis weights \cite{xiao2005scheme} given by
\begin{align*}
c_t(i,j)=&~ \bigl(  1+\max [ d_t(i),d_t(j)]  \bigr)^{-1} , ~~\forall (i,j)\in\mathcal{E}_t, \notag\\
c_t(i,i)=&~1-\sum_{j\in\mathcal N_t(i)}c_t(i,j), ~~\forall i\in\mathcal N,
\end{align*}
where $\mathcal N_t(i) = \{ j\in \mathcal N \colon (j,i) \in \mathcal E_t\} $ is  the set of   neighbors of agent $i$ at time $t$, and 
$d_t(i)=|\mathcal N_t(i)|$ is the degree of agent $i$.

\iffalse  
$$\text{receive } \widetilde{\omega}^j_{t-1} \text{ from neighbors } j \in \mathcal{N}_t(i) \text{over network},$$
$$\omega^i_t = \sum_{j \in \mathcal{N}} c_{t-1}(i,j) \widetilde{\omega}^j_{t-1},$$
$$\text{execute } a^i_t \sim \mu_i(s_t, \cdot),$$
$$\rho^i_t = \frac{\pi^i_{\theta^i_t}(s_t, a^i_t)}{\mu_i(s_t,a^i_t)},$$
$$p^i_t = \log \rho^i_t,$$
$$\text{observe } a_t, r^i_{t+1}, s_{t+1}.$$
\fi 

Next, the agents enter an inner loop and perform the following, repeating until a consensus average of the original values is achieved. In each iteration of the inner loop,  each agent $i$ broadcasts its local $p_t^i$ to its neighbors and  receives $p_t^j$ from each neighbor $j \in \mathcal{N}_t(i)$.
Agent $i$ then updates its local log importance sampling ratio via 
$
p^i_t \leftarrow \sum_{j \in \mathcal{N}} c_t(i,j) p^j_t.
$ 
Such an iteration is repeated until consensus is reached, and all the agents break out of the inner loop. 
For directed graphs, the average consensus can be achieved by using the idea of the push-sum protocol \cite{kempe}; see \cite{ACC12} for a detailed description of the algorithm.

\iffalse 
$$\text{broadcast } p^i_t \text{ and receive } p^j_t \text{ from neighbors } j \in \mathcal{N}_t(i),$$
$$p^i_t \leftarrow \sum_{j \in \mathcal{N}} c_t(i,j) p^j_t.$$
\fi

After achieving consensus, we now have $p^i_t = p^j_t$ for all $i, j \in \mathcal{N}$. Notice that $p^i_t = \frac{1}{n} \sum_{i=1}^n \log \rho^i_t,$ so that $\exp (n p^i_t) = \exp(\sum_{i=1}^n \log \rho^i_t) = \prod_{i=1}^n \rho^i_t = \rho_t = \pi_{\theta_t}(a_t|s_t) / \mu(a_t|s_t)$.

Each agent then performs the local critic and actor updates. For the critic update, agent $i$ first computes the emphatic weighting and the importance sampling ratio \begin{align*}
    M_t = \lambda + (1-\lambda) F_t, \qquad  \rho_t = \exp (n p^i_t),
\end{align*}
where $F_t $ is given by 
$ F_t = 1 + \gamma \rho_{t-1} F_{t-1}$. Notice that this update will be identical across agents. Then, agent $i$ updates its critic parameter $ \widetilde{\omega}^i_t $ via  
\begin{align*}
    e_t &  = \gamma \lambda e_{t-1} + M_t \nabla_{\omega} v_{\omega^i_t}(s_t),\\
    \widetilde{\omega}^i_t & = \omega^i_t + \beta_{\omega,t} \rho_t \delta^i_t e_t,
\end{align*}
where $\delta^i_t = r^i_{t+1} + \gamma v_{\omega^i_t}(s_{t+1}) - v_{\omega^i_t}(s_t)$ is the TD-error computed locally by agent $i$. The parameter $\widetilde{\omega}^i_t $ is then broadcast to all the neighbors in $\mathcal{N}_t(i)$. 
 Finally, for the actor update, the emphatic weighting $M_t^\theta$ is obtained by 
 $$
 M^{\theta}_t = 1 + \lambda^{\theta} \gamma \rho_{t-1} F_{t-1},
 $$
 and the parameter of the local policy $\pi_{\theta}^i$ is updated via 
 \begin{align*}
     \theta^i_{t+1} = \theta^i_t + \beta_{\theta,t} \rho_t M^{\theta}_t \nabla_{\theta^i} \log \pi^i_{\theta^i_t} (s_t, a^i_t) \delta^i_t.
 \end{align*}
For a concise presentation of the algorithm, we refer to the Algorithm 1 pseudocode in the appendix.

\iffalse 
$$\rho_t = \exp (n p^i_t),$$
$$F_t = 1 + \gamma \rho_{t-1} F_{t-1},$$
$$M_t = \lambda + (1-\lambda) F_t,$$
$$e_t = \rho_t(\gamma \lambda e_{t-1} + M_t \nabla_{\omega} v_{\omega^i_t}(s_t)),$$
$$\widetilde{\omega}^i_t = \omega^i_t + \beta_{\omega,t}(r^i_{t+1} + \gamma v_{\omega^i_t}(s_{t+1}) - v_{\omega^i_t}(s_t)) e_t,$$
and finally the actor update, where we set $\delta^i_t = r^i_{t+1} + \gamma v_{\omega^i_t}(s_{t+1}) - v_{\omega^i_t}(s_t)$:
$$M^{\theta}_t = 1 + \lambda^{\theta} \gamma \rho_{t-1} F_{t-1},$$
$$\theta^i_{t+1} = \theta^i_t + \beta_{\theta,t} \rho_t M^{\theta}_t \nabla_{\theta^i} \log \pi^i_{\theta^i_t} (s_t, a^i_t) \delta^i_t,$$
$$\text{broadcast } \widetilde{\omega}^i_t \text{ to neighbors over network}.$$
\fi

\subsection*{Inner Loop Issue}

The inner consensus loop used to obtain the importance sampling ratios in the multi-agent case is a disadvantage of our algorithm. It is important to remark that the unrealistic assumption that the inner loop is run until convergence at every step is for purposes of the theoretical analysis. Two options for relaxing this theoretical assumption while obtaining similar theoretical results are as follows. First, by taking the diameter of the communication graph into consideration, it is likely possible to carefully bound the approximation errors accumulated by running a predetermined, finite number of inner loop steps at each iteration, and use these bounds to prove a modified version of Theorem \ref{convergence:critic} below. Second, it may be possible to use a function approximator $\rho_{\eta} : S \times A \rightarrow [0, \infty)$ parametrized by $\eta \subset \rn{L}, L \in \mathbb{N}, L > 0$ to approximate the global importance sampling ratio, updating its parameters $\eta$ on a third, fastest timescale. In practice, the number of consensus steps needed for the inner loop to provide a nearly unbiased estimate of the true importance sampling ratio will likely be finite and depend in a predictable way on the diameter of the communication graph, allowing the user to simply truncate the inner loop appropriately. It also seems likely that in practice we can perform only a small number of consensus steps at each inner loop with negligible effects on the overall performance of the algorithm -- we briefly discuss this possibility in our experimental results Section \ref{results:empirical}. In conclusion, though we use the importance sampling loop in the current form of the algorithm, it can likely be overcome in both theory and practice. Determining exactly how this will be achieved is an important future direction.

\section{Theoretical Results} \label{results:theoretical}

The general structure of the multi-agent algorithm is that of a two-timescale stochastic approximation scheme with the $\omega$-updates occurring at the faster timescale and the $\theta$-updates occurring at the slower one. As is standard in such schemes \cite{borkar}, in our convergence analysis we first prove the a.s. convergence of the faster timescale updates while viewing the slower timescale $\theta$ and corresponding policy $\pi_{\theta}$ as static, and then show the a.s. convergence of the $\theta$ updates while viewing the value of the faster timescale $\omega$ as equilibrated at every timestep. In our case, of course, we have the additional complications that experience is being generated by each agent $i$ according to a fixed behavior policy $\mu^i$, the critic updates are achieved using a multi-agent, consensus-based version of ETD$(\lambda)$, and we are using an off-policy gradient sampling scheme in our actor updates, but our convergence analysis still follows this two-stage pattern: Theorem \ref{convergence:critic} provides convergence of the critic updates, while Theorem \ref{convergence:actor} provides convergence of the actor updates.

\subsection{Assumptions} \label{assumptions}
The following is a list of the assumptions needed in the proofs of Theorems \ref{convergence:critic} and \ref{convergence:actor}. Assumptions \ref{a:1}, \ref{a:2}, and \ref{a:3}  are taken directly from \cite{zhang}. \ref{a:4} is a standard condition in stochastic approximation. \ref{a:5} requires that the behavior policy be sufficiently exploratory, and also allows us to bound the importance sampling ratios $\rho_t$, which is critical in our convergence proofs. \ref{a:6} simplifies the convergence analysis in the present work, but, as mentioned above, the assumption can likely be weakened or removed by carefully bounding the errors resulting from terminating the inner loop after a specified level of precision is achieved.

\vspace{.1in}

\noindent
\begin{assumption} \label{a:1}
For each agent $i \in \mathcal{N}$, the local $\theta$-update is carried out using the projection operator $\Gamma^i : \rn{m_i} \rightarrow \Theta^i \subset \rn{m_i}$. Furthermore, the set $\Theta = \prod_{i=1}^n \Theta^i$ contains at least one local optimum of $J_{\mu}(\theta)$.
\end{assumption}
\begin{assumption} \label{a:2}
For each element $C_t \in \{C_t\}_{t \in \mathbb{N}}$, 
\begin{enumerate}
\item $C_t$ is row stochastic, $E[C_t]$ is column stochastic, and there exists $\alpha \in (0,1)$ such that, for any $c_t(i,j) > 0$, we have $c_t(i,j) \geq \alpha$.
\item If $(i,j) \notin \mathcal{E}_t$, we have $c_t(i,j) = 0$.
\item The spectral norm $\rho = \rho(E[C^T_t (I - \bm{1} \bm{1}^T / N)C_t])$ satisfies $\rho < 1$.
\item Given the $\sigma$-algebra $\sigma(C_{\tau}, \{r^i_{\tau}\}_{i \in \mathcal{N}}; \tau \leq t)$, $C_t$ is conditionally independent of $r^i_{t+1}$ for each $i \in \mathcal{N}$.
\end{enumerate}
\end{assumption}
\begin{assumption} \label{a:3}
The feature matrix $\Phi$ has linearly independent columns, and the value function approximator $v_{\omega}(s) = \phi(s)^T \omega$ is linear in $\omega$.
\end{assumption}
\begin{assumption} \label{a:4}
We have $\sum_t \beta_{\omega,t} = \sum_t \beta_{\theta,t} = \infty, \ \sum_t \beta_{\omega,t}^2 + \beta_{\theta,t}^2 < \infty, \ \beta_{\theta,t} = o(\beta_{\omega,t})$, and $\lim_{t \rightarrow \infty} \frac{\beta_{\omega,t+1}}{\beta_{\omega,t}} = 1$.
\end{assumption}
\begin{assumption} \label{a:5}
For some fixed $0 < \varepsilon \leq \frac{1}{|S|\cdot |A|}$, we have $\varepsilon \leq \mu(a|s)$, for all state-action pairs $(s,a) \in S \times A$.
\end{assumption}
\begin{assumption} \label{a:6}
Each agent performs its update at timestep $t$ using the exact value of $\rho_t$.
\end{assumption}

\subsection{Convergence} \label{convergence}

As indicated above, our theoretical analysis proceeds in two stages. For the first step of the analysis we prove that, for a fixed target policy $\pi_{\theta}$ and behavior policy $\mu$, when using linear function approximation the multi-agent version of ETD($\lambda$) given in the critic step of our algorithm converges in the following sense: almost surely, each agent asymptotically obtains a copy of the unique solution $\omega_{\theta} \stackrel{\text{def}}{=} \omega^* = -C^{-1} b$ described in Section \ref{etd}, which provides each agent with the best approximator $\Phi \omega_{\theta}$ of the global value function $v_{\theta}$ for the multi-agent MDP under policy $\pi_{\theta}$. More concisely, we have the following:

\begin{theorem} \label{convergence:critic}
Given a fixed target policy $\pi_{\theta}$ and behavior policy $\mu$, multi-agent ETD($\lambda$) achieves consensus a.s. when using linear function approximation, and, under Assumption \ref{a:3}, the consensus vector is a.s. the unique solution of (\ref{etd:eq1}).
\end{theorem}

\noindent See Appendix \ref{appendix:a} for the proof.

For the second step of our analysis, we show that the vector
$\theta_t = 
\begin{bmatrix}
	(\theta^1_t)^T & \ldots & (\theta^n_t)^T 
\end{bmatrix}^T$
of the agents' policy parameters converges a.s. to an equilibrium point $\theta^*$ of a certain ODE ((\ref{actor:eq1}) given below). Let
\eqnn{A^i_t = r^i_{t+1} + \gamma \phi^T_{t+1} \omega^i_t - \phi^T_t \omega^i_t, \hspace{2mm} \psi^i_t = \nabla_{\theta^i} \log \pi^i_{\theta^i_t} (a_t|s_t),}
and $\mathcal{G}_t = \sigma(\theta_{\tau}; \tau \leq t)$ be the $\sigma$-algebra generated by the $\theta$-iterates up to time $t$. Define
\eqnn{A^i_{t,\theta} = r^i_{t+1} + \gamma \phi^T_{t+1} \omega_{\theta} - \phi^T_t \omega_{\theta},}
where $\omega_{\theta}$ is the limit of the critic step at the faster timestep under target policy $\pi_{\theta}$. We then have the following:
\begin{theorem} \label{convergence:actor}
The update
\eqn{actor:eq1}{\theta^i_{t+1} = \Gamma^i(\theta^i_t + \beta_{\theta,t} \rho_t M^{\theta}_t A^i_t \psi^i_t)}
converges a.s. to the set of asymptotically stable equilibria of the ODE
\eqn{actor:eq2}{\dot{\theta}^i = \hat{\Gamma}^i (h^i(\theta)),}
where $h^i(\theta_t) = \CE{\rho_t M^{\theta}_t A^i_{t,\theta_t} \psi^i_t}{\mathcal{G}_t}$ and $\hat{\Gamma}$ is defined as in (\ref{gamma:hat}).
\end{theorem}
\noindent For the proof we again refer the reader to the appendix.

Theorem \ref{convergence:actor} is in the same vein as the classic convergence results for single-agent actor-critic under linear function approximation architectures \cite{bhatnagar09, bhatnagar10, degris12}. It is important to note that, since the approximation $\Phi \omega_{\theta}$ obtained during the critic step is in general a biased estimate of the true value function $v_{\theta}$, the term $\rho_t M^{\theta}_t A^i_{t, \theta} \phi^i_t$ will usually also be a biased estimate of the true policy gradient. However, given that the error between the true value function $v_{\theta^*}$ and the estimate $\Phi \omega_{\theta^*}$ is small, the point $\theta^*$ will lie within a small neighborhood of a local optimum of (\ref{off_objective}), as noted in \cite{zhang}.

\section{Empirical Evaluation} \label{results:empirical}

To illustrate the theoretical results obtained in this paper, we evaluate the multi-agent algorithm with linear value function approximation, as in Assumption \ref{a:3}. Similar to the setup in \cite{zhang}, we consider a multi-agent MDP with simplified transition dynamics: $N = 10$ agents, a state space $S$ with $|S| = 20$, binary actions $A^i = \{ 0, 1 \}$ for each agent $i$, and a transition probability matrix $P$ on $S$ whose entries are drawn uniformly from the interval $[0, 1]$ and normalized to be stochastic. For each agent $i$, the value of its reward function at $(s, a^i)$ is specified at initialization by sampling uniformly from the interval $[0,4]$ for each element $(s, a^i) \in S \times A^i$. The feature vectors $\phi(s) \in \rn{K}$ are generated uniformly from $[0,1]$, for each $s \in S$, where $K = 10$, and the matrix $\Phi$ whose rows are made up of these $\phi(s)$ is verified to have full column rank. A fixed consensus weight matrix $C$ is obtained by randomly generating a connected communication graph $G$ over the agents. Decreasing stepsizes $\beta_{\omega, t} = 1/t^{0.6}, \beta_{\theta, t} = 1 / t^{0.85}$ are chosen, satisfying Assumption \ref{a:4}.

For each target policy $\pi^i_{\theta^i}$ we use the softmax policy
\begin{equation}
    \pi^i_{\theta^i} (s, a^i) = \frac{e^{h_{\theta^i}(s, a^i)}}{\sum_{a^i \in A^i} e^{h_{\theta^i}(s, a^i)}},
\end{equation}
where $h^i_{\theta^i} S \times A^i \rightarrow \r$ is a single-layer neural network with 64 hidden units, sigmoid activation function $f(x) = (1 + e^{-x})^{-1}$ for the hidden layer, and linear activation for the output layer, satisfying the differentiability condition on the target policies assumed in Section \ref{model}. Finally, we assume that the behavior policy $\mu^i$ is the uniform distribution over $A^i$.

\begin{figure}[!htbp]
\begin{subfigure}{.5\textwidth}
  \centering
  \includegraphics[width=0.9\linewidth]{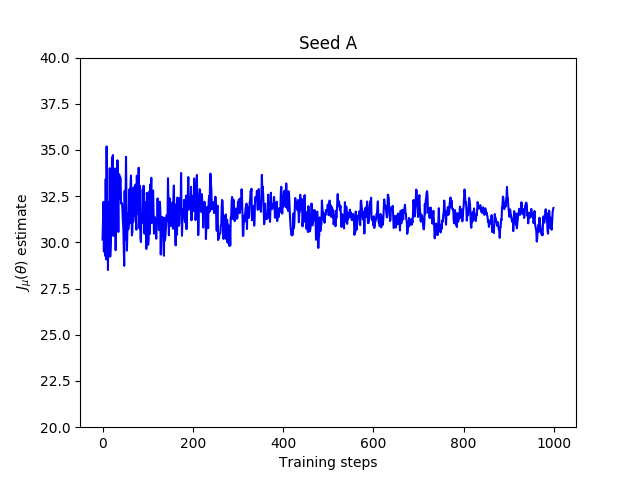}
  \label{fig:sfig1}
\end{subfigure}
\begin{subfigure}{.5\textwidth}
  \centering
  \includegraphics[width=0.9\linewidth]{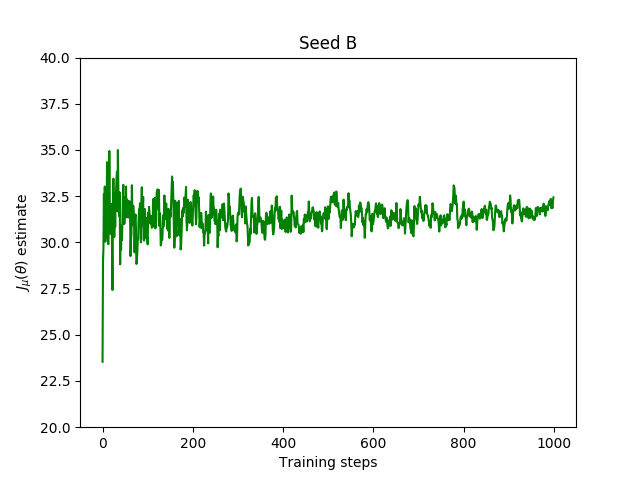}
  \label{fig:sfig2}
\end{subfigure}
\begin{subfigure}{.5\textwidth}
  \centering
  \includegraphics[width=0.9\linewidth]{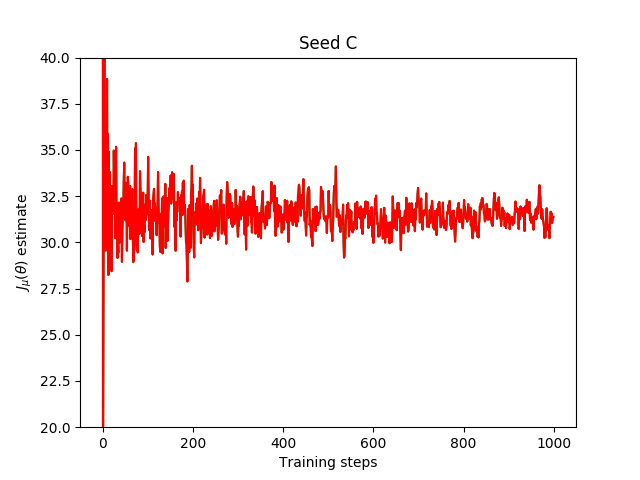}
  \label{fig:sfig2}
\end{subfigure}
\begin{subfigure}{.5\textwidth}
  \centering
  \includegraphics[width=0.9\linewidth]{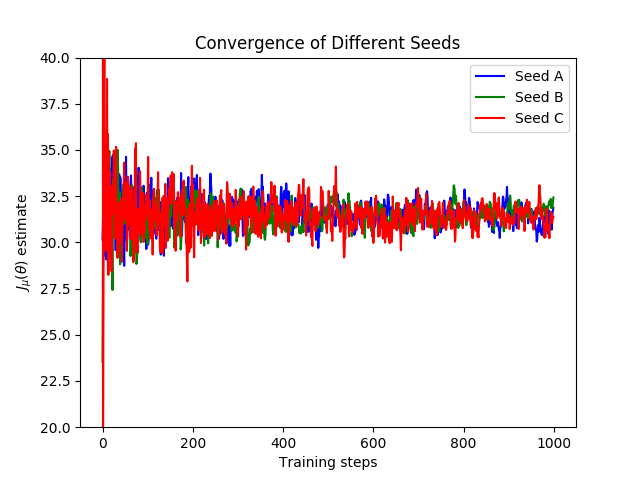}
  \label{fig:sfig2}
\end{subfigure}
\caption{Convergence Under Linear Value Function Approximation}
\label{fig:fig}
\end{figure}

After generating a single problem instance using the procedure described above, we ran several instances of the algorithm on it, where each of the instances used a different random seed to generate the agents' initial parameters. To illustrate convergence, for three of these instances we plot the off-policy objective (\ref{off_objective}) as a function of the number of update steps carried out using the behavior policy. For each seed, the algorithm clearly converges to approximately the same objective function value. That all three seeds converge to the same value may seem surprising given that nonlinear function approximation is used in the policy, but the problem considered here is quite simply, so this should not be expected, in general. In these figures, at each step the agents effectively carry out the inner consensus loop until convergence. However, when we ran the same experiments on the same seeds in two additional configurations -- (i) with only three steps of the inner consensus loop carried out, and (ii) with no inner consensus loop (i.e. each agent uses only its own $\rho_t^i$) -- we found the performance to be almost exactly the same. This suggests that the inner consensus loop issue can be overcome in practice, though further experiments on more complex problems are warranted.

\newpage

\section{Conclusions}

In this paper, we have extended off-policy actor-critic methods to the multi-agent reinforcement learning context. In order to accomplish this, we first extended emphatic temporal difference learning to the multi-agent setting, which allowed us to perform policy evaluation during the critic step. We then provided a novel multi-agent off-policy policy gradient theorem, which gave access to the policy gradient estimates needed for the actor step. With these tools at hand, we proposed a new multi-agent off-policy actor-critic algorithm and proved its convergence when linear function approximation of the state-value function is used. We also provided empirical results to both back up our theoretical claims and motivate further study of our algorithm. Based on the foundations provided in this paper, promising future directions of research include exploration of additional theoretical applications of multi-agent emphatic temporal difference learning, further practical and theoretical exploration of the inner loop issue, empirical comparison of our algorithm with other off-policy multi-agent reinforcement learning algorithms, such as that in \cite{yanzhang19}, and the development of practical applications of our work.

\bibliographystyle{plain}
\bibliography{refs}

\appendix
\section{Appendix} \label{appendix:a}

\subsection{Proof of Theorem \ref{pgt}}

\noindent {\em Proof:}
Following \cite{imani}, we first have that 
$$\nabla_{\theta} J_{\mu}(\theta) = \nabla_{\theta} \sum_{s \in S} d_{\mu}(s) v_{\theta}(s) = \sum_{s \in S} d_{\mu}(s) \nabla_{\theta} v_{\theta}(s),$$
so it suffices to consider $\nabla_{\theta} v_{\theta}(s)$. Now
\begin{eqnarray*}
\nabla_{\theta} v_{\theta} (s) 
&=& \nabla_{\theta} \sum_{a \in A} \pi_{\theta}(a|s)q_{\theta}(s,a) \\
&=& \sum_{a \in A} \Big[ [\nabla_{\theta} \pi_{\theta}(a|s)] q_{\theta}(s,a) + \pi_{\theta}(a|s) \nabla_{\theta}q_{\theta}(s,a) \Big] \\
%
%\small
%$$= \sum_{a \in A} \Big[ [\nabla_{\theta} \pi_{\theta}(s,a)] q_{\theta}(s,a) + \pi_{\theta}(s,a) \nabla_{\theta} \Big[ \sum_{s' \in S} P(s' |  s,a)(\bar{r}(s,a) + \gamma v_{\theta}(s')) \Big] \Big]$$
%
%\small
&=& \sum_{a \in A} \Big[ [\nabla_{\theta} \pi_{\theta}(a|s)] q_{\theta}(s,a) + \gamma \pi_{\theta}(a|s) \sum_{s' \in S} P(s' | s,a) \nabla_{\theta} v_{\theta}(s')) \Big].
\end{eqnarray*}
%\normalsize
%
Letting $\mathbf{V}_{\theta} \in \mathbb{R}^{|S| \times d}$ denote the matrix of gradients $\nabla_{\theta} v_{\theta}(s)$ for each $s \in S$, and $\mathbf{G} \in \mathbb{R}^{|S| \times d}$ the matrix with rows $\mathbf{g}(s)^T$ given by
$$\mathbf{g}(s) = \sum_{a \in A} [\nabla_{\theta} \pi_{\theta}(a|s)] q_{\theta}(s,a),$$
the last expression above can be rewritten as $\mathbf{V}_{\theta} = \mathbf{G} + \mathbf{P}_{\theta, \gamma} \mathbf{V}_{\theta}$, i.e. $\mathbf{V}_{\theta} = (\mathbf{I} - \mathbf{P}_{\theta, \gamma})^{-1} \mathbf{G}$. We thus finally have
\begin{eqnarray*}
\nabla_{\theta} J_{\mu}(\theta) &=& \mathbf{d}_{\mu}^T \mathbf{V}_{\theta} = \mathbf{d}_{\mu}^T (\mathbf{I} - \mathbf{P}_{\theta,\gamma})^{-1} \mathbf{G} = \mathbf{m}^T \mathbf{G}\\
&=&  \sum_{s \in S} m(s) \sum_{a \in A} [\nabla_{\theta} \pi_{\theta}(a|s)] q_{\theta}(s,a).
\end{eqnarray*}
Now notice that, in our multi-agent case,
\begin{eqnarray*}
[\nabla_{\theta} \pi_{\theta}(a|s)] q_{\theta}(s,a) 
&=& \pi_{\theta} (a|s) [\nabla_{\theta} \log \pi_{\theta}(a|s)] q_{\theta}(s,a)\\
&=& \pi_{\theta}(a|s) \Big[ \nabla_{\theta} \log \prod_{i \in \mathcal{N}} \pi^i_{\theta^i}(a^i|s) \Big] q_{\theta}(s,a)\\ 
&=& \pi_{\theta}(a|s) \sum_{i \in \mathcal{N}} [ \nabla_{\theta}\log \pi^i_{\theta^i}(a^i|s)] q_{\theta}(s,a),
\end{eqnarray*}
which implies that
$$\nabla_{\theta^i} J_{\mu}(\theta) = \sum_{s \in S} m(s) \sum_{a \in A} \pi_{\theta}(a|s) q_{\theta}(s,a) \nabla_{\theta^i} \log \pi^i_{\theta^i}(a^i|s),$$
which concludes the proof.
\hfill $\qed$

\subsection{Proof of Theorem \ref{convergence:critic}}

We now prove the convergence of consensus ETD$(\lambda)$.

%%%
With 
$\omega_t = 
\begin{bmatrix}
	(\omega^1_t)^T & \ldots & (\omega^n_t)^T 
\end{bmatrix}^T$
and $e_t$ defined as in Section \ref{etd}, we can write the global $\omega$-update for all agents as
\eqnn{\omega_{t+1} = (C_t \otimes I)(\omega_t + \beta_{\omega,t} \Delta_t),}
where 
\eqnn{\delta_t = \rvec{\delta^1_t \ldots \delta^n_t}^T, \ \delta^i_t = r^i_{t+1} + \gamma \phi_{t+1}^T \omega^i_t - \phi^T_t \omega^i_t, \ \Delta_t = \delta_t \otimes e_t.}
Define
\begin{eqnarray*}
\langle \omega \rangle &=& \frac{1}{n} (\bm{1}^T \otimes I) \omega = \frac{1}{n} \sum_{i \in \mathcal{N}} \omega_i, \\
T(\omega) &=& \bm{1} \otimes \langle \omega \rangle, \\
\omega_{\perp} &=& T_{\perp} (\omega) = \omega - T(\omega) = ((I - \frac{1}{n} \bm{1} \bm{1}^T) \otimes I) \omega.
\end{eqnarray*}
The vector $\langle \omega \rangle$ is simply the average over the $\omega^i$ for all agents $i$. The vector $T(\omega)$ is called the {\em agreement vector}, and $\omega_{\perp}$ the {\em disagreement vector}.
%%%

To prove Theorem \ref{convergence:critic}, we first assume that $\sq{\omega_t}{t \in \n}$ is a.s. bounded, and then show that $\omega_{\perp,t} \rightarrow 0$ a.s., which means that all agents do reach consensus a.s. Under the same assumption, we then prove that $\lim_t \langle \omega_t \rangle = \omega^*$ a.s. Finally, we verify the conditions of Theorem \ref{consensus:stability:theorem} and invoke it to obtain a.s. boundedness of $\sq{\omega_t}{t \in \n}$. Proving Theorem \ref{convergence:critic} thus reduces to proving Lemmas \ref{convergence:critic:lemma:1}, \ref{convergence:critic:lemma:2}, and \ref{convergence:critic:lemma:3} below.
\begin{lemma} \label{convergence:critic:lemma:1}
$\omega_{\perp,t} \rightarrow 0$ a.s.
\begin{proof}
Notice that $\omega_t = T(\omega_t) + \omega_{\perp,t} = \bm{1} \otimes \langle \omega_t \rangle + \omega_{\perp,t}$. This allows us to write
\begin{eqnarray*}
\omega_{\perp, t+1} &=& T_{\perp}(\omega_{t+1}) = T_{\perp} ((C_t \otimes I)(\bm{1} \otimes \langle \omega_t \rangle + \omega_{\perp,t} + \beta_{\omega,t} \rho_t \Delta_t)) \\
&=& T_{\perp}( \bm{1} \otimes \langle \omega_t \rangle + (C_t \otimes I)(\omega_{\perp,t} + \beta_{\omega,t} \rho_t \Delta_t)) \\
&=& T_{\perp} ((C_t \otimes I)(\omega_{\perp,t} + \beta_{\omega,t} \rho_t \Delta_t)) \\
&=& (C_t \otimes I)(\omega_{\perp,t} + \beta_{\omega,t} \rho_t \Delta_t) - \bm{1} \otimes \langle (C_t \otimes I)(\omega_{\perp,t} + \beta_{\omega,t} \rho_t \Delta_t) \rangle \\
&=& (C_t \otimes I)(\omega_{\perp,t} + \beta_{\omega,t} \rho_t \Delta_t) - (\bm{1} \otimes I) \langle (C_t \otimes I)(\omega_{\perp,t} + \beta_{\omega,t} \rho_t \Delta_t) \rangle \\
&=& (C_t \otimes I)(\omega_{\perp,t} + \beta_{\omega,t} \rho_t \Delta_t) - (\bm{1} \otimes I)(\frac{1}{n}(\bm{1}^T \otimes I))(C_t \otimes I)(\omega_{\perp,t} + \beta_{\omega,t} \rho_t \Delta_t) \\
&=& (C_t \otimes I)((I-\frac{1}{n} \bm{1} \bm{1}^T) \otimes I)(\omega_{\perp,t} + \beta_{\omega,t} \rho_t \Delta_t) \\
&=& ((I-\frac{1}{n} \bm{1} \bm{1}^T) \otimes I)(C_t \otimes I)(\omega_{\perp,t} + \beta_{\omega,t} \rho_t \Delta_t).
\end{eqnarray*}

By hypothesis, we have that $P(\sup_t \norm{\omega_t} < \infty) = P(\cup_{M \in \n} \{ \sup_t \norm{z_t} \leq M \}) = 1$, and by property \ref{etd:traces:1} of the trace iterates (see Section \ref{etd:traces}) we similarly have $P(\sup_t \norm{e_t} < \infty) = P(\cup_{M \in \n} \{ \sup_t \norm{e_t} \leq M \}) = 1$. Thus, to prove $\omega_{\perp,t} \rightarrow 0$ a.s., it suffices to show that $\lim_t \omega_{\perp, t} \mathbb{I}_{\{ \sup_t \norm{z_t} \leq M \}} = 0$ for all $M \in \n$, where $\mathbb{I}_{\{\cdot\}}$ is the indicator function and $z_t = (\omega_t, e_t)$.

If we can show that, for any $M \in \n$,
\eqnn{\sup_t E[ \norm{\beta_{\omega,t}^{-1} \omega_{\perp, t}}^2 \mathbb{I}_{\{\sup_t \norm{z_t} \leq M\}}] < \infty,}
this will imply that there exists $K > 0$ such that \eqnn{E[\norm{\omega_{\perp,t}}^2 \mathbb{I}_{\{ \sup_t \norm{z_t} \leq M \}}] \leq K \beta_{\omega,t}^2.}
Summing over both sides yields that $\sum_t E[\norm{\omega_{\perp,t}}^2 \mathbb{I}_{\{ \sup_t \norm{z_t} \leq M\}}]$ is finite, whence $\sum_t \norm{\omega_{\perp,t}}^2 \mathbb{I}_{\{ \sup_t \norm{z_t} \leq M \}}$ is finite a.s., and thus $\lim_t \omega_{\perp,t} \mathbb{I}_{\{ \sup_t \norm{z_t} \leq M \}} = 0$ a.s., as desired.
\\ \\
To demonstrate that
\eqnn{\sup_t E[ \norm{\beta_{\omega,t}^{-1} \omega_{\perp, t}}^2 \mathbb{I}_{\{\sup_t \norm{z_t} \leq M\}}] < \infty,}
we proceed as follows. We first have
\begin{eqnarray*}
&& \norm{\beta^{-1}_{\omega,t+1} \omega_{\perp,t+1}}^2 \\
&=& \beta^{-2}_{\omega,t+1}(\omega_{\perp,t} + \beta_{\omega,t} \rho_t \Delta_{t+1})^T((I - \frac{1}{n} \bm{1}\bm{1}^T)C_t \otimes I)^T ((I- \frac{1}{n}\bm{1}\bm{1}^T)C_t \otimes I)(\omega_{\perp,t} + \beta_{\omega,t} \rho_t \Delta_{t+1}) \\
&=& \beta^{-2}_{\omega,t+1}(\omega_{\perp,t} + \beta_{\omega,t} \rho_t \Delta_{t+1})^T(C_t^T(I - \frac{1}{n} \bm{1}\bm{1}^T)C_t \otimes I)(\omega_{\perp,t} + \beta_{\omega,t} \rho_t \Delta_{t+1}) \\
&=& \frac{\beta^2_{\omega,t}}{\beta^2_{\omega,t+1}}(\beta^{-1}_{\omega_t} \omega_{\perp,t} + \rho_t \Delta_{t+1})^T(C_t^T(I - \frac{1}{n} \bm{1}\bm{1}^T)C_t \otimes I)(\beta^{-1}_{\omega_t} \omega_{\perp,t} + \rho_t \Delta_{t+1}).
\end{eqnarray*}
Recalling parts 3 and 4 of Assumption \ref{a:2}, we have
\begin{eqnarray*}
&& E[\norm{\beta^{-1}_{\omega,t+1} \omega_{\perp,t+1}}^2 \ | \ \mathcal{F}_t] \\ &\leq& \rho \frac{\beta^2_{\omega,t}}{\beta^2_{\omega,t+1}} E[\norm{\beta^{-1}_{\omega_t} \omega_{\perp,t} + \rho_t \Delta_{t+1}}^2 \ | \ \mathcal{F}_t] \\
&\leq& \rho \frac{\beta^2_{\omega,t}}{\beta^2_{\omega,t+1}} \Big[ E[\norm{\beta^{-1}_{\omega,t} \omega_{\perp,t}}^2 \ | \ \mathcal{F}_t] + 2 E[\norm{\beta^{-1}_{\omega,t} \rho_t \Delta_{t+1}^T \omega_{\perp,t}} \ | \ \mathcal{F}_t] + E[\norm{\rho_t \Delta_{t+1}}^2 \ | \ \mathcal{F}_t] \Big] \\
&\leq& \rho \frac{\beta^2_{\omega,t}}{\beta^2_{\omega,t+1}} \Big[ \norm{\beta^{-1}_{\omega,t} \omega_{\perp,t}}^2 + 2 \norm{\beta^{-1}_{\omega,t} \omega_{\perp,t}} E[\norm{\rho_t \Delta_{t+1}}^2 \ | \ \mathcal{F}_t]^{1/2} + E[\norm{\rho_t \Delta_{t+1}}^2 \ | \ \mathcal{F}_t] \Big] \\
&\leq& \rho \frac{\beta^2_{\omega,t}}{\beta^2_{\omega,t+1}} \Big[ \norm{\beta^{-1}_{\omega,t} \omega_{\perp,t}}^2 + \frac{2}{\varepsilon} \norm{\beta^{-1}_{\omega,t} \omega_{\perp,t}} E[\norm{\Delta_{t+1}}^2 \ | \ \mathcal{F}_t]^{1/2} + \frac{1}{\varepsilon^2} E[\norm{\Delta_{t+1}}^2 \ | \ \mathcal{F}_t] \Big],
\end{eqnarray*}
where the third inequality is an application of the Cauchy-Schwarz inequality and the fourth is by \ref{a:5}. The terms containing $\omega_{\perp,t}$ are a.s. bounded, so we just need to bound the terms containing $\Delta_{t+1}$. We have
\begin{eqnarray*}
\norm{\Delta_{t+1}}^2 &=& \norm{\delta \otimes e_t}^2 = \sum_{i \in \mathcal{N}} \norm{(r^i_{t+1} + \gamma \phi^T_{t+1} \omega^i_t - \phi^T_t \omega^i_t)e_t}^2 \\
&=& \sum_{i \in \mathcal{N}} \norm{r^i_{t+1}e_t + ((\gamma \phi^T_{t+1} \omega^i_t - \phi^T_t) \omega^i_t)e_t}^2 \\
&\leq& \sum_{i \in \mathcal{N}} \Big(\norm{r^i_{t+1} e_t}^2 + 2 \norm{r^i_{t+1} e_t} \cdot \norm{((\gamma \phi^T_{t+1} - \phi^T_t)\omega^i_t)e_t} + \norm{((\gamma \phi^T_{t+1} - \phi^T_t)\omega^i_t)e_t}^2 \Big) \\
&=& \sum_{i \in \mathcal{N}} \Big(|r^i_{t+1}|^2 \norm{e_t}^2 + 2 | r^i_{t+1}| \cdot \norm{e_t} \cdot |(\gamma \phi^T_{t+1} - \phi^T_t)\omega^i_t| \cdot \norm{e_t} + |(\gamma \phi^T_{t+1} - \phi^T_t)\omega^i_t|^2 \norm{e_t}^2 \Big) \\
&\leq& \sum_{i \in \mathcal{N}} \Big(|r^i_{t+1}|^2 \norm{e_t}^2 + 2 | r^i_{t+1}| \cdot \norm{\gamma \phi^T_{t+1} - \phi^T_t} \cdot \norm{\omega^i_t} \cdot \norm{e_t}^2 + \norm{\gamma \phi^T_{t+1} - \phi^T_t}^2 \norm{\omega^i_t}^2 \norm{e_t}^2 \Big).
\end{eqnarray*}
Since the state and action spaces are finite, the rewards $r^i_{t+1}$ and feature vectors $\phi_{t+1}, \phi_t$ are bounded. So, for any $M > 0$, there exists $K_1 > 0$ such that $E[\norm{\Delta_{t+1}}^2 \ | \ \mathcal{F}_t] < K_1$ on the set $\{ \sup_{\tau \leq t} \norm{z_{\tau}} \leq M \}$, i.e.
\eqnn{E[\norm{\Delta_{t+1}}^2 \mathbb{I}_{\{ \sup_{\tau \leq t} \norm{z_{\tau}} \leq M \}} \ | \ \mathcal{F}_t ] \leq K_1.}

Now, noticing that $\mathbb{I}_{\{ \sup_{\tau \leq t+1} \norm{z_{\tau}} \leq M \}} \leq \mathbb{I}_{\{ \sup_{\tau \leq t} \norm{z_{\tau}} \leq M \}}$, we combine this bound with the preceding one to get
\begin{eqnarray*}
&& E[\norm{\beta^{-1}_{\omega,t+1} \omega_{\perp,t+1}}^2 \mathbb{I}_{\{ \sup_{\tau \leq t+1} \norm{z_{\tau}} \leq M \}} \ | \ \mathcal{F}_t] \\
&\leq& \rho \frac{\beta^2_{\omega,t}}{\beta^2_{\omega,t+1}} \Big[ \norm{\beta^{-1}_{\omega,t} \omega_{\perp,t}}^2 \mathbb{I}_{\{ \sup_{\tau \leq t} \norm{z_{\tau}} \leq M \}} \\
&& + \frac{2}{\varepsilon} \norm{\beta^{-1}_{\omega,t} \omega_{\perp,t}} \mathbb{I}_{\{ \sup_{\tau \leq t} \norm{z_{\tau}} \leq M \}} E[\norm{\Delta_{t+1}}^2 \mathbb{I}_{\{ \sup_{\tau \leq t} \norm{z_{\tau}} \leq M \}} \ | \ \mathcal{F}_t]^{1/2} \\
&& + \frac{1}{\varepsilon^2} E[\norm{\Delta_{t+1}}^2 \mathbb{I}_{\{ \sup_{\tau \leq t} \norm{z_{\tau}} \leq M \}} \ | \ \mathcal{F}_t] \Big] \\
&\leq& \rho \frac{\beta^2_{\omega,t}}{\beta^2_{\omega,t+1}} \Big[ \norm{\beta^{-1}_{\omega,t} \omega_{\perp,t}}^2 \mathbb{I}_{\{ \sup_{\tau \leq t} \norm{z_{\tau}} \leq M \}} + \frac{2}{\varepsilon} \sqrt{K_1} \cdot \norm{\beta^{-1}_{\omega,t} \omega_{\perp,t}} \mathbb{I}_{\{ \sup_{\tau \leq t} \norm{z_{\tau}} \leq M \}}  + \frac{1}{\varepsilon^2} K_1 \Big].
\end{eqnarray*}
Let $\kappa_t = \norm{\beta^{-1}_{\omega,t} \omega_{\perp,t}}^2 \mathbb{I}_{\{ \sup_{\tau \leq t} \norm{z_{\tau}} \leq M \}}$. Recalling the double expectation formula and taking expectations gives
\begin{eqnarray*}
E[\kappa_{t+1}] &\leq& \rho \frac{\beta^2_{\omega,t}}{\beta^2_{\omega,t+1}} \Big[ E[\kappa_t] + \frac{2}{\varepsilon} \sqrt{K_1} E[\sqrt{\kappa_t}] + \frac{1}{\varepsilon} K_1 \Big] \\
&\leq& \rho \frac{\beta^2_{\omega,t}}{\beta^2_{\omega,t+1}} \Big[ E[\kappa_t] + \frac{2}{\varepsilon} \sqrt{K_1} \sqrt{E[\kappa_t]} + \frac{1}{\varepsilon} K_1 \Big],
\end{eqnarray*}
where the last is by Jensen's inequality. Since $\rho \in [0,1)$ and $\lim_t \frac{\beta_{\omega,t}}{\beta_{\omega,t+1}} = 1$, for any $\delta \in (0,1)$ we may choose $t_0$ such that $\rho \frac{\beta^2_{\omega,t}}{\beta^2_{\omega,t+1}} < 1-\delta$, for all $t \geq t_0$. Then, for $t \geq t_0$,
\eqnn{E[\kappa_{t+1}] \leq (1-\delta) \Big[ E[\kappa_t] + \frac{2}{\varepsilon} \sqrt{K_1} \sqrt{E[\kappa_t]} + \frac{1}{\varepsilon^2} K_1 \Big].}
There furthermore exist $K_2, K_3 > 0$ such that
\eqnn{E[\kappa_{t+1}] \leq (1-\delta) \Big[ E[\kappa_t] + \frac{2}{\varepsilon} \sqrt{K_1} \sqrt{E[\kappa_t]} + \frac{1}{\varepsilon} K_1 \Big] \leq (1-\frac{\delta}{2}) E[\kappa_t] + K_2 \mathbb{I}_{\{E[\kappa_t] < K_3\}}.}
Expanding this gives $E[\kappa_t] \leq (1-\delta/2)^{t-t_0} E[\kappa_{t_0}] + 2K_2 / \delta,$ for $t \geq t_0$, whence $\sup_t E[\kappa_t] < \infty$. Since $\mathbb{I}_{\{ \sup_t \norm{z_t} \leq M \}} \leq \mathbb{I}_{\{ \sup_{\tau \leq t} \norm{z_{\tau}} \leq M \}}$, we finally have
\eqnn{\sup_t E[\norm{\beta^{-1}_{\omega,t} \omega_{\perp,t}}^2 \mathbb{I}_{\{ \sup_t \norm{z_t} \leq M \}}] < \infty.}
\end{proof}
\end{lemma}

In order to prove the following lemma, we manipulate the $\langle \omega \rangle$-update into a form that we recognize as tracking the mean ODE
\eqn{critic:eq1}{\dot{\omega} = C \omega + b}
of the ETD($\lambda$) updates associated with the projected generalized Bellman equation
\eqn{critic:eq2}{v = \Pi(\bar{r}^{\lambda}_{\pi_{\theta},\gamma} + P^{\lambda}_{\pi,\gamma}v).}
We prove that the stochastic approximation conditions hold, implying that these updates almost surely converge to the unique solution $\omega_{\theta} = -C^{-1}b$ such that $\Phi \omega_{\theta}$ solves the above projected equation.

\begin{lemma} \label{convergence:critic:lemma:2}
$\lim_t \langle \omega_t \rangle = \omega_{\theta}$ a.s.
\begin{proof}
Consider the update equation
\begin{eqnarray*}
\langle \omega_{t+1} \rangle &=& \frac{1}{n}(\bm{1}^T \otimes I)(C_t \otimes I)(\bm{1} \otimes \langle \omega_t \rangle + \omega_{\perp,t} + \beta_{\omega,t} \rho_t \Delta_t) \\
&=& \langle \omega_t \rangle + \beta_{\omega,t} \langle (C_t \otimes I)(\rho_t \Delta_t + \beta^{-1}_{\omega,t} \omega_{\perp,t}) \rangle.
\end{eqnarray*}
Rewriting, we can express the update as
\eqn{critic:eq3}{\langle \omega_{t+1} \rangle = \langle \omega_t \rangle + \beta_{\omega,t} E[\rho_t e_t \langle \delta_t \rangle \ | \ \mathcal{F}_t] + \xi_{t+1}),}
where
\eqnn{\xi_{t+1} = \langle (C_t \otimes I)(\rho_t \Delta_t + \beta^{-1}_{\omega,t} \omega_{\perp,t}) \rangle - E[\rho_t e_t \langle \delta_t \rangle \ | \ \mathcal{F}_t].}
Update (\ref{critic:eq3}) has mean ODE (\ref{critic:eq1}). We clearly have that $h(\omega_t) = E[\rho_t e_t \langle \delta_t \rangle \ | \mathcal{F}_t]$ is Lipschitz continuous in $\omega_t$. Since $\{ \langle \omega_t \rangle \}$ is a.s. bounded by assumption, and $\sum_t \beta_{\omega,t} = \infty, \ \sum_t \beta^2_{\omega,t} < \infty$, we only need to verify that $\{ \xi_t \}$ is a martingale difference sequence satisfying
\eqn{critic:eq4}{E[\norm{\xi_{t+1}}^2 \ | \ \mathcal{F}_t] \leq K(1 + \norm{\omega_t}^2)}
for some $K > 0$.

By \ref{a:2}, $E[C_t]$ is doubly stochastic and conditionally independent of $\langle \delta_t \rangle$, whence
\begin{eqnarray*}
&& E[ \langle (C_t \otimes I) (\rho_t \Delta_t + \beta^{-1}_{\omega,t} \omega_{\perp,t}) \rangle \ | \ \mathcal{F}_t] \\
&=& E[ \langle (C_t \otimes I) \rho_t \Delta_t \rangle \ | \ \mathcal{F}_t] = E[\frac{1}{n}(\bm{1}^T \otimes I)(C_t \otimes I) \rho_t \Delta_t \ | \ \mathcal{F}_t] \\
&=& E[\frac{1}{n}(\bm{1}^T \otimes I) \rho_t \Delta_t \ | \ \mathcal{F}_t] = E[\rho_t e_t \langle \delta_t \rangle \ | \ \mathcal{F}_t],
\end{eqnarray*}
since $\langle \omega_{\perp,t} \rangle = 0$ and 
\begin{eqnarray*}
\langle \rho_t \Delta_t \rangle &=& \frac{1}{n} \sum_{i \in \mathcal{N}} \rho_t (r^i_{t+1} + \gamma \phi^T_{t+1} \omega^i_t - \phi^T_t \omega^i_t) e_t \\
&=& \rho_t e_t (\bar{r}^i_{t+1} + (\gamma \phi^T_{t+1} + \phi^T_t) \langle \omega_t \rangle) = \rho_t e_t \langle \delta_t \rangle.
\end{eqnarray*}
$\xi_{t+1}$ is thus a martingale difference sequence. To see that (\ref{critic:eq4}) is satisfied, first note that
\eqn{critic:eq5}{\norm{\xi_{t+1}}^2 \leq 3 \norm{\frac{1}{n}(\bm{1}^T \otimes I)(C_t \otimes I)(\rho_t \Delta_{t+1} + \beta^{-1}_{\omega,t} \omega_{\perp,t})}^2 + 3 \norm{E[\rho_t e_t \langle \delta \rangle \ | \ \mathcal{F}_t]}^2.}
Considering the first term in (\ref{critic:eq5}), we have
\begin{eqnarray*}
&& \norm{\frac{1}{n}(\bm{1}^T \otimes I)(C_t \otimes I)(\rho_t \Delta_{t+1} + \beta^{-1}_{\omega,t} \omega_{\perp,t})}^2 \\
&=& (\rho_t \Delta_{t+1} + \beta^{-1}_{\omega,t} \omega_{\perp,t})( C^T_t \bm{1}\bm{1}^T C_t \otimes \frac{1}{n^2} I)(\rho_t \Delta_{t+1} + \beta^{-1}_{\omega,t} \omega_{\perp,t}).
\end{eqnarray*}
Since $C_t$ is doubly stochastic in expectation, the matrix $( C^T_t \bm{1}\bm{1}^T C_t \otimes \frac{1}{n^2} I)$ has spectral norm that is bounded in expectation, so we may choose $K_4 > 0$ such that
\eqnn{E[\norm{\frac{1}{n}(\bm{1}^T \otimes I)(C_t \otimes I)(\rho_t \Delta_{t+1} + \beta^{-1}_{\omega,t} \omega_{\perp,t})}^2 \ | \ \mathcal{F}_t]
\leq K_4 E[\norm{(\rho_t \Delta_{t+1} + \beta^{-1}_{\omega,t} \omega_{\perp,t})}^2 \ | \ \mathcal{F}_t].}
By Cauchy-Schwarz, our proof for \ref{convergence:critic:lemma:1}, and the a.s. boundedness of $\{ \omega_t \}$, we can further choose $K_5 > 0$ such that the above is
\eqnn{\leq K_4 E \Big[ \frac{2}{\varepsilon^2} \norm{\Delta_{t+1}}^2 + 2 \norm{\beta^{-1}_{\omega,t} \omega_{\perp,t}}^2 \ | \ \mathcal{F}_t \Big] \leq K_5.}
Consider now the rightmost term in (\ref{critic:eq5}). Recall that $\rho_t \leq \frac{1}{\varepsilon}$, for all $t \geq 0$. Choose $K_6 > 0$ such that $\sup_t \norm{e_t} < \frac{\varepsilon}{\sqrt{2}} \sqrt{K_6}$ a.s. Then,
\begin{eqnarray*}
2 \norm{E[\rho_t e_t \langle \delta_t \rangle \ | \ \mathcal{F}_t ]}^2 &\leq& K_6 \norm{E[\langle \delta_t \rangle \ | \ \mathcal{F}_t]}^2 = K_6 \norm{E[\overline{r}_{t+1} + (\gamma \phi^T_{t+1} - \phi^T_t) \langle \omega_t \rangle \ | \ \mathcal{F}_t]}^2 \\
&=& K_6 \norm{E[\overline{r}_{t+1} \ | \ \mathcal{F}_t] + E[(\gamma \phi^T_{t+1} - \phi^T_t) \langle \omega_t \rangle \ | \ \mathcal{F}_t]}^2 \leq K_6 (K_7 + K_8 \norm{\langle \omega_t \rangle}^2) \\
&\leq& K_9(1 + \norm{\langle \omega_t \rangle}^2),
\end{eqnarray*}
for some constants $K_7, K_8, K_9 > 0$, where the second-to-last inequality follows from an application of Cauchy-Schwarz, Jensen's inequality, and Cauchy-Schwarz again.
\end{proof}
\end{lemma}

All that remains to prove now is the a.s. boundedness of $\{ \omega_t \}$. We do so by verifying the conditions of Theorem \ref{consensus:stability:theorem}.

\begin{lemma} \label{convergence:critic:lemma:3}
$\sup_t \norm{\omega_t} < \infty$ a.s.
\begin{proof}
We can write the consensus update for agent $i$ as
\eqnn{\omega^i_{t+1} = \sum_{j \in \mathcal{N}} c_t(i,j)[\omega^j_t + \beta_{\omega,t} \rho_t e_t \delta^j_t] = \sum_{j \in \mathcal{N}} c_t(i,j)[\omega^j_t + \beta_{\omega,t} (h^j(\omega_t,Z_t) + \xi^j_t)],}
where $h^j(\omega_t,Z_t) = \CE{\rho_t e_t \delta^j_t}{\mathcal{F}_t},$ and $\xi^j_t = \rho_t e_t \delta^j_t - \CE{\rho_t e_t \delta^j_t}{\mathcal{F}_t}$.

The first two conditions of Theorem \ref{consensus:stability:theorem} are easily verified. For \ref{b:1}, to see that $h^j$ is continuous in its first argument, fix $Z \in S \times A \times \rn{k} \times \r$, and $\omega_1, \omega_2 \in \rn{kn}$. We have by the boundedness of the rewards $r^j_{t+1}$ and feature vectors $\phi_t, \phi_{t+1}$ that
\eqnn{\norm{h^j(\omega_1,Z) - h^j(\omega_2,Z)} = \norm{\CE{\rho_t e_t (\gamma\phi^T_{t+1} - \phi^T_t)(\omega^j_1 - \omega^j_2)}{\mathcal{F}_t}} \leq K_{10}\norm{\omega^j_1 - \omega^j_2}
\leq K_{10}\norm{\omega_1 - \omega_2},}
for some $K_{10} > 0$, whence $h^j$ is Lipschitz continuous in $\omega$. For \ref{b:2}, the sequence $\xi_t = [(\xi^1_t)^T \ldots (\xi^n_t)^T]^T$ is clearly a martingale, and an argument analogous to that used to prove condition (\ref{critic:eq4}) in \ref{convergence:critic:lemma:2} can be used to show
\eqnn{\CE{\norm{\xi^j_{t+1}}^2}{\mathcal{F}_t} \leq K_{11} (1 + \norm{\omega^j_t}^2)}
for some $K_{11} > 0,$ which in turn implies the existence of $K_{12} > 0$ such that
\eqnn{\CE{\norm{\xi_{t+1}}^2}{\mathcal{F}_t} \leq K_{12} (1 + \norm{\omega_t}^2).}

Verifying condition \ref{b:3} of Theorem \ref{consensus:stability:theorem} is less straightforward. Let $\zeta_{t+1} = \overline{h}(\omega_t) - h(\omega_t,Z_t)$, where $\overline{h}^i(\omega_t) = E_{\eta}[h^i(\omega_t,Z_t)]$, where $\eta$ is the unique invariant probability measure associated with the Markov chain $\sq{Z_t}{t \in \n}$. We need to show that there exists $K > 0$ such that $\norm{\zeta_{t+1}}^2 \leq K(1 + \norm{\omega_t}^2)$ a.s. It suffices to prove there exists $K > 0$ such that
\eqn{critic:eq6}{\norm{\zeta^i_{t+1}}^2 \leq K(1 + \norm{\omega^i_t}^2) \text{ a.s.}}
First note that
\eqnn{\norm{\overline{h}^i(\omega_t) - h^i(\omega_t,Z_t)}^2 \leq 3 \norm{\overline{h}^i(\omega_t)}^2 + 3 \norm{h^i(\omega_t,Z_t)}^2.}
Considering the first term we obtain
\begin{eqnarray*}
\norm{\overline{h}^i(\omega_t)}^2 &=& \norm{\Es{\eta}{\CE{\rho_t e_t(r^i_{t+1} + \gamma \phi^T_{t+1} \omega^i_t - \phi^T_t \omega^i_t)}{\omega_t,C_{t-1}}}}^2 \\
&=& \norm{\Es{\eta}{\CE{\rho_t e_t r^i_{t+1}}{\omega_t, C_{t-1}} + \CE{(\rho_t(\gamma \phi^T_{t+1} - \phi^T_t) \omega^i_t) e_t}{\omega_t, C_{t-1}}}}^2 \\
&\leq& K_{13} \norm{\Es{\eta}{e_t}}^2 + K_{14} \norm{\Es{\eta}{\CE{((\gamma \phi^T_{t+1} - \phi^T_t) \omega^i_t) e_t}{\omega_t, C_{t-1}}}}^2 \\
&\leq& K_{14} \Es{\eta}{\norm{\CE{((\gamma \phi^T_{t+1} - \phi^T_t) \omega^i_t) e_t}{\omega_t, C_{t-1}}}^2} + K_{15} \\
&\leq& K_{14} \Es{\eta}{\CE{\norm{((\gamma \phi^T_{t+1} - \phi^T_t) \omega^i_t) e_t}^2}{\omega_t, C_{t-1}}} + K_{15} \\
&=& K_{14} \Es{\eta}{\CE{|((\gamma \phi^T_{t+1} - \phi^T_t) \omega^i_t|^2 \norm{e_t}^2}{\omega_t, C_{t-1}}} + K_{15} \\
&\leq& K_{14} \Es{\eta}{\CE{\norm{\gamma \phi^T_{t+1} - \phi^T_t}^2 \norm{\omega^i_t}^2 \norm{e_t}^2}{\omega_t, C_{t-1}}} + K_{15} \\
&\leq& K_{16} \Es{\eta}{\norm{e_t}^2 \CE{\norm{\omega^i_t}^2}{\omega_t, C_{t-1}}} + K_{15} \\
&\leq& K_{17}(1 + \CE{\norm{\omega^i_t}^2}{\omega_t}) = K_{17}(1 + \norm{\omega^i_t}^2),
\end{eqnarray*}
for some $K_{13}, \ldots, K_{17} > 0$. The second inequality follows from Jensen's inequality and the a.s. boundedness of $\{e_t\}$. The third follows from Jensen's inequality and the fact that, for real-valued random variables $X, Y$ satisfying $0 \leq X \leq Y$ a.s., we have $\E{X} \leq \E{Y}$. The fourth inequality is an application of Cauchy-Schwarz. The fifth follows from the boundedness of the feature vectors. The final inequality follows from the a.s. boundedness of $\{e_t\}$ and the fact that the integrand is independent of $C_{t-1}$.

A similar argument shows that, in light of the a.s. boundedness of $\{e_t\}$, there exists $K > 0$ such that $\norm{h^i(\omega_t,Z_t)}^2 \leq K(1 + \norm{\omega^i_t}^2)$ a.s., which proves (\ref{critic:eq6}).

Let $h_c, \overline{h},$ and $\widetilde{h}_c$ be defined as in Theorem \ref{consensus:stability:theorem}, and $C$ and $b$ as in (\ref{etd:eq1}). We complete the proof of the current lemma by verifying \ref{b:4}. For $c > 0$ and $x \in \rn{k}$, we have
\eqnn{\widetilde{h}_c(x) = Cx + \frac{1}{c} b.}
As $c \rightarrow \infty,$ and fixing $x \in K$, where $K$ is any compact set, we have that $\lim_{c \rightarrow \infty} h_c(x) = h_{\infty}(x)$ exists and $h_{\infty}(x) = Cx$. The ODE $\dot{x} = h_{\infty}(x)$ clearly has 0 as its globally asymptotically stable attractor, which completes the proof.
\end{proof}

\end{lemma}

\subsection{Proof of Theorem \ref{convergence:actor}}
We now prove convergence of the actor step.

\begin{proof}
Rewrite the update (\ref{actor:eq1}) as follows:
\eqn{actor:eq3}{\theta^i_{t+1} = \Gamma^i(\theta^i_t + \beta_{\theta,t}(h^i(\theta^i_t) + \zeta^i_{t,1} + \zeta^i_{t,2})),}
where
\eqnn{\zeta^i_{t,1} = \rho_t M^{\theta}_t A^i_t \psi^i_t - \CE{\rho_t M^{\theta}_t A^i_t \psi^i_t}{\mathcal{G}_t}, \hspace{2mm} \zeta^i_{t,2} = \CE{\rho_t M^{\theta}_t (A^i_t - A^i_{t,\theta_t}) \psi^i_t}{\mathcal{G}_t}.}

To prove the theorem, we would like to be able to apply the Kushner-Clark Lemma, presented as Theorem \ref{kc:lemma} in the appendix, for (\ref{actor:eq3}). Before that theorem applies, however, we need to demonstrate that $h^i$ is continuous in $\theta$. To see that this is indeed the case, it suffices to show that the integrand $\rho_t M^{\theta}_t A^i_{t,\theta_t} \psi^i_t$ is continuous in $\theta_t$.

Since $\pi_{\theta}$ is assumed to be continuously differentiable and $\theta_t$ is restricted to lie in a compact set, we have that $\rho_t \psi^i_t$ is continuous in $\theta_t$. Second, $M^{\theta}_t$ is a finite sum of products of functions continuous in $\theta_t$, so it too is continous in $\theta_t$. Third, $\pi_{\theta}$ is continuous and the transition probabilities $P(s' \ | \ s, a)$ are given for each $(s, a) \in S \times A$, which yields that the entries of both $P^{\lambda}_{\gamma,\pi_{\theta_t}}$ and $r^{\lambda}_{\gamma, \pi_{\theta_t}}$ are continuous functions of $\theta_t$. This implies that $C$ and $b$ from (\ref{etd:eq1}) are continous in $\theta_t$, whence $\omega_{\theta_t} = -C^{-1} b$ is continuous in $\theta_t$, and thus $h^i$ is continuous in $\theta_t$.

Condition \ref{kc:1} of Theorem \ref{kc:lemma} is satisfied by hypothesis, and condition \ref{kc:3} follows from the proof of the critic step, since $\omega_t \rightarrow \omega_{\theta}$ a.s. and thus $A^i_t \rightarrow A^i_{t,\theta_t}$ a.s., so it remains to verify condition \ref{kc:2}.

Notice that, since $\theta_t$ is restricted to lie in a compact set, and $\rho_t \psi^i_t, M^{\theta}_t,$ and $A^i_t$ are continuous in $\theta_t$, we have that $\sq{\zeta_{t,1}}{t \in \n}$ is a.s. bounded, so
\eqnn{\sum_t \norm{\beta_{\theta,t} \zeta_{t+1,1}}^2 < \infty \text{ a.s.}}
Define $\mathcal{M}_t = \sum_{\tau=0}^t \beta_{\theta,t} \zeta_{t+1,1}$, for each $t \in \n$. Clearly $\sq{\mathcal{M}_t}{t \in \n}$ is a martingale. By the above, however, we also have that
\eqnn{\sum_t \norm{\mathcal{M}_{t+1} - \mathcal{M}_t}^2 = \sum_t \norm{\beta_{\theta,t} \zeta_{t+1,1}}^2 < \infty \text{ a.s.},}
so $\sq{\mathcal{M}_t}{t \in \n}$ converges a.s. by the martingale convergence theorem. This means that
\eqnn{\lim_t \Big( \sup_{n \geq t} \norm{\sum_{\tau = t}^n \beta_{\theta,\tau} \zeta_{\tau+1,1}} \geq \epsilon \Big) = 0,}
for all $\epsilon > 0$, which completes the verification of Theorem \ref{kc:lemma} and thus the proof.

\end{proof}

\section{Previously Existing Results} \label{appendix:b}

In this section we collect together for reference some important results from the literature that we have used in the development of our algorithm and our convergence proofs.

\subsection{ETD($\lambda$) Trace Iterates} \label{etd:traces}
From \cite{yu15} we have the following important properties concerning the trace iterates $\{(e_t, F_t)\}_{t \in \mathbb{N}}$. Letting $Z_t = (s_t,a_t,e_t,F_t)$, for $t \in \mathbb{N}$, we have the following:

\begin{condition} \label{etd:traces:1}
$\{Z_t\}_{t \in \mathbb{N}}$ is an ergodic Markov chain with a unique invariant probability measure $\eta$.
\end{condition}
\begin{condition} \label{etd:traces:2}
For any initial $(e_0,F_0)$, $\sup_{t \in \mathbb{N}} E[\norm{(e_t,F_t)}] < \infty$.
\end{condition}
\noindent Note that \ref{etd:traces:1} implies that $\{e_t\}_{t \in \mathbb{N}}$ is a.s. bounded.

\subsection{Stability of the Consensus Updates} \label{consensus:stability}
To prove a.s. boundedness of the critic updates $\{\omega^i_t\}_{t \in \n}$, we rely on the following slight generalization of a theorem proven in the appendix of \cite{zhang}. The consensus update for agent $i$ can be expressed as
\eqn{stability:update}{\omega^i_{t+1} = \sum_{j \in \mathcal{N}} c_t(i,j) [\omega^j_t + \beta_{\omega,t} (h^j(\omega_t, Z_t) + \xi^j_{t+1})],}
where $\omega_t = 
\begin{bmatrix}
	(\omega^1_t)^T & \ldots & (\omega^n_t)^T 
\end{bmatrix}^T$,
$Z_t$ is the Markov chain with unique invariant probability measure $\eta$ associated with the trace iterates, $c_t(i,j) = [C_t]_{ij}$, $h^j$ is an $\rn{n}$-valued function, and $\sq{\xi^j_t}{t \in \n}$ is a martingale difference sequence with respect to $\sq{\mathcal{F}_t}{t \in \n}$ defined below. Note that, in (\ref{stability:update}), the function $h^j(\omega_t,Z_t)$ depends only on $(\omega^j_t, Z_t)$ in our context.

For the following, let
$\overline{h}^i(\omega_t) = E_{\eta}[h^i(\omega_t,Z_t)]$,  
$h=\rvec{(h^1)^T \ldots(h^n)^T}^T$, 
$\overline{h}  = \rvec{(\overline{h}^1)^T \ldots (\overline{h}^n)^T}^T$, and
$\xi_t = \rvec{(\xi^1_t)^T \ldots (\xi^n_t)^T}^T$.
Let $\sq{\mathcal{F}_t}{t \in \n}$ be the filtration defined by $\mathcal{F}_t = \sigma(\omega_{\tau}, Z_{\tau}, C_{\tau - 1}; \tau \leq t)$. Define $h_c : \rn{kn} \rightarrow \rn{kn}$ by $h_c(\omega) = c^{-1}\overline{h}(c \omega)$ for $c > 0$, and $\widetilde{h}_c(x) : \rn{k} \rightarrow \rn{k}$ by $\widetilde{h}(x) = \langle h_c(\bm{1} \otimes x) \rangle$, where $\otimes$ is the Kronecker product and $\langle \cdot \rangle : \rn{kn} \rightarrow \rn{k}$ is given by
\eqnn{\langle \omega \rangle = \frac{1}{n} (\bm{1}^T \otimes I) \omega = \frac{1}{n} \sum_{i \in \mathcal{N}} \omega_i.}
Consider the following conditions:

\begin{condition} \label{b:1}
$h^i : \rn{kn} \times S \times A \times \rn{k} \times \r \rightarrow \rn{k}$ is Lipschitz continuous in its first argument $\omega \in \rn{kn}$, for all $i \in \mathcal{N}$.
\end{condition}
\begin{condition} \label{b:2}
The martingale difference sequence $\sq{\xi_t}{t \in \n}$ satisfies
\eqnn{E[\norm{\xi_{t+1}}^2 \ | \ \mathcal{F}_t] \leq K ( 1 + \norm{\omega_t}^2)}
for some $K>0$.
\end{condition}
\begin{condition} \label{b:3}
The difference $\zeta_{t+1} = \overline{h}(\omega_t) - h(\omega_t,Z_t)$ satisfies
\eqnn{\norm{\zeta_{t+1}}^2 \leq K' (1 + \norm{\omega_t}^2)}
a.s., for some $K'>0$.
\end{condition}
\begin{condition} \label{b:4}
There exists $h_{\infty} : \rn{k} \rightarrow \rn{k}$ such that, as $c \rightarrow \infty$, $\widetilde{h}_c(x)$ converges uniformly to $h_{\infty}(x)$ on compact sets, and, for some $\epsilon < 1/\sqrt{n}$, the set $\set{x}{\norm{x} \leq \epsilon}$ contains a globally aymptotically stable attractor of the ODE
\eqnn{\dot{x} = h_{\infty}(x).}
\end{condition}
\noindent We then have:
\begin{theorem} \label{consensus:stability:theorem}
Under assumptions \ref{a:2}, \ref{a:4}, \ref{b:1}, \ref{b:2}, \ref{b:3}, and \ref{b:4}, the sequence $\sq{\omega_t}{t \in \n}$ is a.s. bounded.
\end{theorem}
\noindent The proof can be found in \cite{zhang}. The original statement of the theorem in that paper required that the Markov chain $\sq{Z_t}{t \in \n}$ have a finite state space. This assumption is in fact unnecessary, so long as \ref{b:3} is still satisfied.

\subsection{Stochastic Approximation} \label{sa}
The following classical stochastic approximation conditions taken from \cite{borkar}. Consider the stochastic approximation scheme in $\rn{k}$ given by the update equation
\eqn{sa:eq1}{x_{n+1} = x_n + \alpha_n [h(x_n) + \mathcal{M}_{n+1}],}
where $n \in \n$ and $x_0$ is given. Consider also the following assumptions.
\begin{condition} \label{sa:1}
$h : \rn{k} \rightarrow \rn{k}$ is Lipschitz continuous.
\end{condition}
\begin{condition} \label{sa:2}
$\sq{\alpha_n}{n \in \n}$ satisfies $\sum_n \alpha_n = \infty$, $\sum_n \alpha_n^2 < \infty$, and $\alpha_n \geq 0$ for all $n \in \n$.
\end{condition}
\begin{condition} \label{sa:3}
$\sq{\mathcal{M}_n}{n \in \n}$ is a martingale difference sequence with respect to the filtration given by $\mathcal{F}_n = \sigma(x_m, \mathcal{M}_m; m \leq n) = \sigma(x_0, \mathcal{M}_m; m \leq n)$, and furthermore
\eqn{sa:eq2}{\CE{\norm{\mathcal{M}_{n+1}}^2}{\mathcal{F}_n} \leq K(1 + \norm{x_n}^2) \text{ a.s.},}
for all $n \in \n$.
\end{condition}
\begin{condition} \label{sa:4}
$\sup_n \norm{x_n} < \infty \text{ a.s.}$
\end{condition}
\noindent Under conditions 1 through 4 above, we have the following theorem.
\begin{theorem} \label{sa:theorem}
The sequence $\sq{x_n}{n \in \n}$ converges a.s. to the set of asymptotically stable equilibria of the ODE
\eqn{sa:eq3}{\dot{x}(t) = h(x(t)), t \geq 0.}
\end{theorem}
\noindent Note that, if (\ref{sa:eq3}) has a unique globally asymptotically stable equilibrium point $x^*$, we have $x_n \rightarrow x^*$ a.s.

\subsection{Kushner-Clark Lemma} \label{kc}
Our convergence result for the actor step relies on the Kushner-Clark lemma \cite{kushner}, which we now state. Let $\Gamma : \rn{k} \rightarrow \rn{k}$ be a projection onto a compact set $K \subset \rn{k}$. Let
\eqn{gamma:hat}{\hat{\Gamma}(h(x)) = \lim_{\epsilon \downarrow 0} \frac{\Gamma(x + \epsilon h(x)) - x}{\epsilon}}
for $x \in K$ and $h : \rn{k} \rightarrow \rn{k}$ continuous on $K$. Consider the update
\eqn{kclemma:eq1}{x_{t+1} = \Gamma(x_t + \alpha_t (h(x_t) + \zeta_{t,1} + \zeta_{t,2}))}
and its associated ODE
\eqn{kclemma:eq2}{\dot{x} = \hat{\Gamma}(h(x)).}
Consider the following three conditions:
\begin{condition} \label{kc:1}
$\sq{\alpha_t}{t \in \n}$ satisfies $\alpha_t, \sum_t \alpha_t = \infty, \sum_t \alpha^2_t < \infty.$
\end{condition}
\begin{condition} \label{kc:2}
$\sq{\zeta_{t,1}}{t \in \n}$ is such that
\eqnn{\lim_t P \Big( \sup_{n \geq t} \norm{\sum_{\tau = t}^n \alpha_{\tau} \zeta_{\tau,1}} \geq \epsilon \Big) = 0,}
for all $\epsilon > 0$.
\end{condition}
\begin{condition} \label{kc:3}
$\sq{\zeta_{t,2}}{t \in \n}$ is an a.s. bounded random sequence with $\zeta_{t,2} \rightarrow 0$ a.s.
\end{condition}

\begin{theorem} \label{kc:lemma}
Under \ref{kc:1}, \ref{kc:2}, and \ref{kc:3}, if (\ref{kclemma:eq2}) has a compact set $K'$ as its asymptotically stable equilibria, then the updates (\ref{kclemma:eq1}) converge a.s. to $K'$.
\end{theorem}

\newpage

\begin{algorithm}
\caption{Multi-agent Off-policy Actor-critic}
Initialize $\theta^i_0 = 0, \omega_0 = e_{-1} = 0, F_{-1} = 0, \rho_{-1} = 1,$ for all $i \in \mathcal{N}$, the initial state $s_0$, and the stepsizes $\sq{\beta_{\omega,t}}{t \in \n}, \sq{\beta_{\theta,t}}{t \in \n}$.
\begin{algorithmic} 
    \Repeat
    \ForAll{$i \in \mathcal{N}$}
        \State $\text{receive } \widetilde{\omega}^j_{t-1} \text{ from neighbors } j \in \mathcal{N}_t(i) \text{ over network}$
        \State $\omega^i_t = \sum_{j \in \mathcal{N}} c_{t-1}(i,j) \widetilde{\omega}^j_{t-1}$
        \State $\text{execute } a^i_t \sim \mu_i(\cdot|s_t)$
        \State $\rho^i_t = \frac{\pi^i_{\theta^i_t}(a^i_t|s_t)}{\mu_i(a^i_t|s_t)}$
        \State $p^i_t = \log \rho^i_t$
        \State $\text{observe } r^i_{t+1}, s_{t+1}$
        \Repeat
        \Comment{begin inner consensus loop}
            \State $\text{broadcast } p^i_t \text{ and receive } p^j_t \text{ from neighbors } j \in \mathcal{N}_t(i)$
            \State $p^i_t \leftarrow \sum_{j \in \mathcal{N}} c_t(i,j) p^j_t$
        \Until{consensus is achieved}
        \Comment{end inner consensus loop}
        \State $\rho_t = \exp (n p^i_t)$
        \State $F_t = 1 + \gamma \rho_{t-1} F_{t-1}$ 
        \Comment{begin critic update}
        \State $M_t = \lambda + (1-\lambda) F_t$
        \State $e_t = \gamma \lambda e_{t-1} + M_t \nabla_{\omega} v_{\omega^i_t}(s_t)$
        \State $\delta^i_t = r^i_{t+1} + \gamma v_{\omega^i_t}(s_{t+1}) - v_{\omega^i_t}(s_t)$
        \State $\widetilde{\omega}^i_t = \omega^i_t + \beta_{\omega,t} \rho_t \delta^i_t e_t$
        \Comment{end critic update}
        \State $M^{\theta}_t = 1 + \lambda^{\theta} \gamma \rho_{t-1} F_{t-1}$
        \Comment{begin actor update}
        \State $\theta^i_{t+1} = \theta^i_t + \beta_{\theta,t} \rho_t M^{\theta}_t \nabla_{\theta^i} \log \pi^i_{\theta^i_t} (a^i_t|s_t) \delta^i_t$
        \Comment{end actor update}
        \State $\text{broadcast } \widetilde{\omega}^i_t \text{ to neighbors over network}$
    \EndFor
    \Until{convergence}
\end{algorithmic}
\end{algorithm}

\end{document}